%% file: main.tex
\documentclass[11pt]{article}
\usepackage[utf8]{inputenc}

\usepackage{microtype}
\usepackage{graphicx}
\usepackage{adjustbox}
\usepackage{subcaption}
\usepackage{booktabs}
\usepackage{microsoft-tech-report}
\usepackage{tabularx}
\usepackage{hyperref}
\usepackage{amsmath}
\usepackage{amssymb}
\usepackage{mathtools}

\usepackage{amsthm}
\usepackage{bm}
\usepackage{booktabs}
\usepackage{float}
\usepackage{multirow}
\usepackage{enumitem}
\usepackage[capitalize,noabbrev]{cleveref}
\usepackage{xspace}
\usepackage{titletoc}
\usepackage{placeins}  
\usepackage{tcolorbox} 
\tcbuselibrary{breakable,skins}
\usepackage{listings}  
\usepackage{hyperref}
\usepackage{marvosym}
\usepackage{adjustbox}
\usepackage{xcolor}
\usepackage{colortbl}

\definecolor{bestblue}{HTML}{B3CDE3}  
\definecolor{secondblue}{HTML}{E0ECF4}

\tcbuselibrary{theorems,skins,breakable,listings}
\definecolor{promptbg}{HTML}{F8F8F8}
\definecolor{promptborder}{HTML}{1F4E79}
\newtcolorbox{promptbox}[1][]{
  colback=promptbg, colframe=promptborder, fonttitle=\bfseries\sffamily,
  breakable, enhanced, arc=3pt, boxrule=0.8pt,
  left=10pt, right=10pt, top=8pt, bottom=8pt,
  title={#1}
}

\lstdefinestyle{json}{
  basicstyle=\small\ttfamily,
  breaklines=true,
  breakatwhitespace=false,
  showstringspaces=false,
  columns=fullflexible,
  literate={"}{\textquotedbl}1,
}

\newif\ifappendixtoc
\appendixtoctrue   

\definecolor{color_blue}{HTML}{E7EFFA}
\definecolor{color_green}{HTML}{E6F8E0}
\definecolor{color_gray}{HTML}{ECECEC}
\definecolor{pearDark}{HTML}{2980B9}
\definecolor{theoremblue}{HTML}{EBF5FB}
\definecolor{theoremborder}{HTML}{2980B9}
\definecolor{propgreen}{HTML}{EAFAF1}
\definecolor{propborder}{HTML}{27AE60}
\definecolor{defyellow}{HTML}{FEF9E7}
\definecolor{defborder}{HTML}{F39C12}
\definecolor{remarkgray}{HTML}{F2F3F4}
\definecolor{remarkborder}{HTML}{7F8C8D}

\hypersetup{
  colorlinks=true,
  linkcolor=msftblue,
  citecolor=msftblue,
  urlcolor=msftblue,
  pdftitle={Mage-VL-Cowork: A Full Training Recipe to Learn Hybrid Computer Use from Evolved Harnesses}
}

\newtcbtheorem[auto counter]{theorem}{Theorem}{
  colback=theoremblue, colframe=theoremborder, fonttitle=\bfseries,
  breakable, enhanced, sharp corners, boxrule=0.6pt, left=6pt, right=6pt, top=6pt, bottom=6pt,
  crefname={Theorem}{Theorems}
}{thm}

\newtcbtheorem[use counter from=theorem]{proposition}{Proposition}{
  colback=propgreen, colframe=propborder, fonttitle=\bfseries,
  breakable, enhanced, sharp corners, boxrule=0.6pt, left=6pt, right=6pt, top=6pt, bottom=6pt,
  crefname={Proposition}{Propositions}
}{prop}

\newtcbtheorem[use counter from=theorem]{lemma}{Lemma}{
  colback=theoremblue, colframe=theoremborder, fonttitle=\bfseries,
  breakable, enhanced, sharp corners, boxrule=0.6pt, left=6pt, right=6pt, top=6pt, bottom=6pt,
  crefname={Lemma}{Lemmas}
}{lem}

\newtcbtheorem[use counter from=theorem]{corollary}{Corollary}{
  colback=propgreen, colframe=propborder, fonttitle=\bfseries,
  breakable, enhanced, sharp corners, boxrule=0.6pt, left=6pt, right=6pt, top=6pt, bottom=6pt,
  crefname={Corollary}{Corollaries}
}{cor}

\newtcbtheorem[use counter from=theorem]{definition}{Definition}{
  colback=defyellow, colframe=defborder, fonttitle=\bfseries,
  breakable, enhanced, sharp corners, boxrule=0.6pt, left=6pt, right=6pt, top=6pt, bottom=6pt,
  crefname={Definition}{Definitions}
}{def}

\newtcbtheorem[use counter from=theorem]{assumption}{Assumption}{
  colback=defyellow, colframe=defborder, fonttitle=\bfseries,
  breakable, enhanced, sharp corners, boxrule=0.6pt, left=6pt, right=6pt, top=6pt, bottom=6pt,
  crefname={Assumption}{Assumptions}
}{asm}

\newtcbtheorem[use counter from=theorem]{remark}{Remark}{
  colback=remarkgray, colframe=remarkborder, fonttitle=\bfseries,
  breakable, enhanced, sharp corners, boxrule=0.6pt, left=6pt, right=6pt, top=6pt, bottom=6pt,
  crefname={Remark}{Remarks}
}{rem}

\crefname{tcb@cnt@theorem}{Theorem}{Theorems}
\crefname{tcb@cnt@proposition}{Proposition}{Propositions}
\crefname{tcb@cnt@lemma}{Lemma}{Lemmas}
\crefname{tcb@cnt@corollary}{Corollary}{Corollaries}
\crefname{tcb@cnt@definition}{Definition}{Definitions}
\crefname{tcb@cnt@assumption}{Assumption}{Assumptions}
\crefname{tcb@cnt@remark}{Remark}{Remarks}
\crefname{algocf}{Algorithm}{Algorithms}

\newcommand{\magevl}{Mage-VL\xspace}

\newcommand{\sysname}{Mage-VL-Cowork\xspace}

\newcommand{\genname}{Mage-CUA-Forge\xspace}

\newif\ifrevmark
\revmarkfalse
\definecolor{revcolor}{HTML}{B45309}
\definecolor{revbg}{HTML}{FEF3C7}
\definecolor{revborder}{HTML}{F59E0B}
\newcommand{\rev}[1]{\ifrevmark{\color{revcolor}#1}\else#1\fi}
\newenvironment{revblock}{\ifrevmark\begin{tcolorbox}[breakable,enhanced,colback=revbg!45,coltext=revcolor,colframe=revborder,boxrule=0pt,borderline west={2.2pt}{0pt}{revborder},sharp corners,left=6pt,right=6pt,top=4pt,bottom=4pt,before skip=4pt,after skip=4pt]\fi}{\ifrevmark\end{tcolorbox}\fi}

\techreportshorttitle{Mage-VL-Cowork}

\begin{document}
\thispagestyle{empty}

\noindent
\begin{minipage}[c]{0.5\linewidth}
\raggedright
\raisebox{-0.5\height}{\msftbrandmark}
\end{minipage}
\begin{minipage}[c]{0.49\linewidth}
\raggedleft
{\msftdatefont\small\color{msftgray}October, 2026}
\end{minipage}\par
\vspace{0.35em}
\noindent{\color{msftline}\rule{\linewidth}{0.8pt}\par}

\vspace{1.0em}
\begin{center}
{{\msfttitlefont\fontsize{21}{25}\selectfont\color{msftdark}
Mage-VL-Cowork: A Full Training Recipe \\ to Learn Hybrid Computer Use From \\ Evolved Harnesses \par}}
\vspace{1.25em}

{\normalsize\rmfamily\color{msftdark}
\hyperref[sec:contributor]{Microsoft Mage Team}\par

}

\end{center}

\vspace{0.35em}
\begin{msfttitlebox}
\setlength{\parindent}{0cm}
\setlength{\parskip}{0cm}

\input{sections/0_abstract}

\vspace{0.14cm}
{\setlength{\parskip}{0.06cm}\small
{\msftmetalabel{Project Page}\href{https://microsoft.github.io/Mage}{https://microsoft.github.io/Mage}\par}
{\msftmetalabel{Code}\href{https://github.com/microsoft/Mage}{https://github.com/microsoft/Mage}\par}
{\msftmetalabel{Model}\href{https://huggingface.co/collections/microsoft/mage}{https://huggingface.co/collections/microsoft/mage}\par}
{\msftmetalabel{Date}October, 2026\par}
}

\end{msfttitlebox}

\vspace{-0.2em}
\begin{figure}[H]
    \centering
    \includegraphics[width=\linewidth]{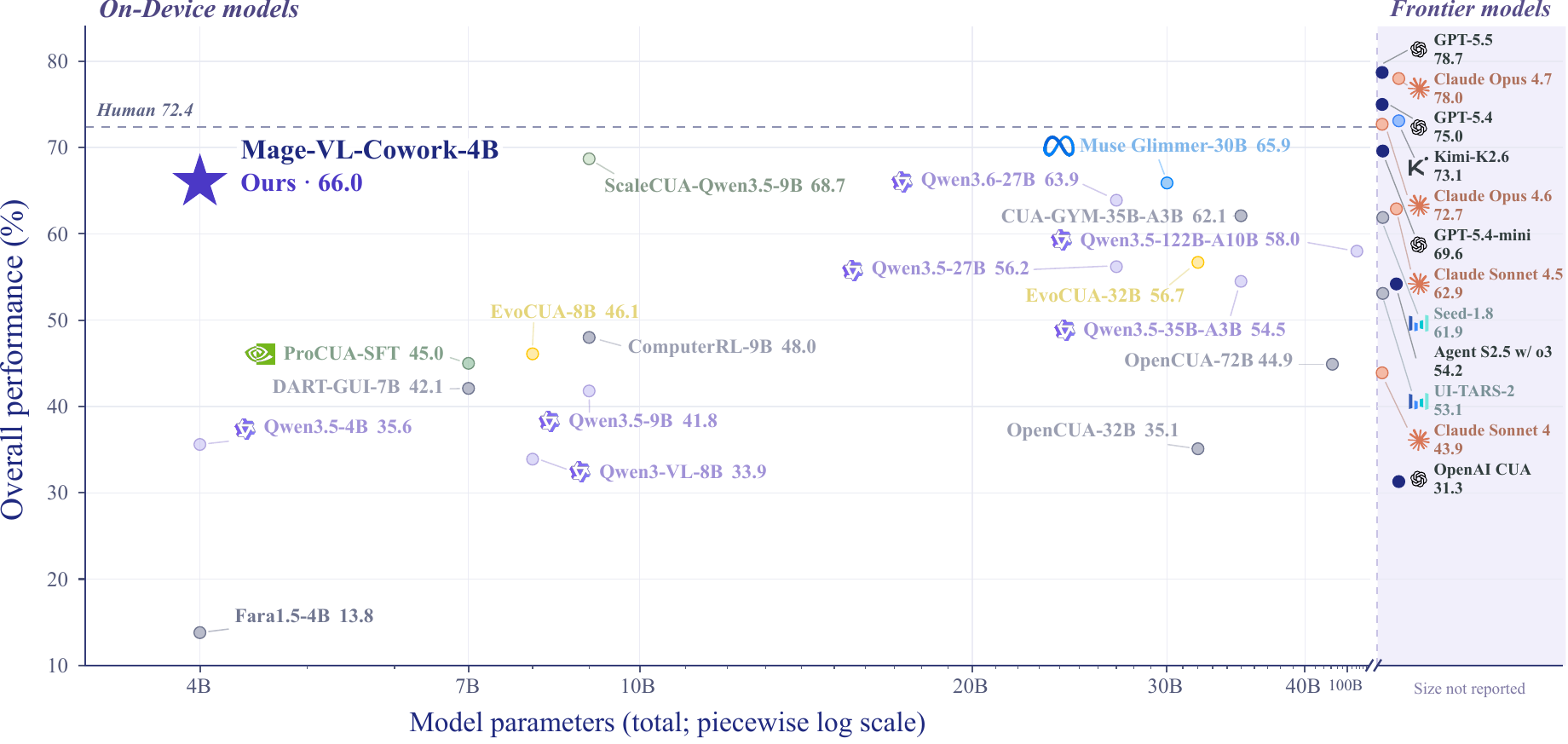}
    \caption{OSWorld-Verified overall score versus model size. Mage-VL-Cowork-4B
    is compared with compact and larger open-source agents; frontier models
    without reported parameter counts are shown in the right strip. The dashed
    line marks human performance (72.4\%).}
    \label{fig:teaser}
\end{figure}

\clearpage
\setcounter{tocdepth}{2}
{\small\tableofcontents}
\clearpage


\input{sections/1_introduction}
\input{sections/2_related_work}
\input{sections/3_datagen_0921}
\input{sections/4_trainpipeline}
\FloatBarrier
\input{sections/4a_rl_infrastructure}
\FloatBarrier
\input{sections/5_experiments}
\input{sections/6_conclusion}
\input{sections/7-author}

\input{sections/appendix}

\newpage
\FloatBarrier  
\bibliography{references,references_additions}
\bibliographystyle{unsrtnat}

\end{document}

%% file: sections/0_abstract.tex
\begin{abstract}
Effective computer-use agents (CUAs) should seamlessly combine visual GUI
interaction with code and terminal execution within a single policy. We present
\sysname{}, a 4B hybrid CUA trained from a general multimodally aligned
checkpoint through an end-to-end, progressive five-stage curriculum across
midtraining, SFT, and RL. To power this curriculum at scale, \genname{}
replaces per-task agent construction with reusable, skill-conditioned
generation workflows and application-specific preparers, synthesizing 293,806
desktop and web tasks across 6,915 skills while reducing recurring per-task
expenditure by 82.6\%. Rather than directly distilling a default teacher, we
evolve the teacher harness and instantiate specialized CUA harnesses to yield
high-quality trajectories alongside stage-specific supervision. \sysname{}-4B
finally climbs to 66.0\% on OSWorld-Verified through different training stages
with more efficient reasoning cost, achieving state-of-the-art performance at
its parameter size and pushing forward the frontier of on-device intelligence.
\end{abstract}

%% file: sections/1_introduction.tex
\section{Introduction}
\label{sec:introduction}

Computer-use agents (CUAs) for productivity perceive graphical desktops through screenshots and
act through keyboard and mouse commands. This GUI channel is universal, but it is a
modeling choice rather than a property of the domain, and an expensive one: 
a typical screenshot can require thousands of tokens, increasing the cost of
processing high-resolution visual input and grounding interface elements.
A person fluent with a computer does not confine
themselves to the mouse: they open a terminal when a terminal is faster and read
a file to confirm what a dialog claims to have saved. 
Restricting the policy to
pixels forces it to re-derive through the GUI what a single command would
accomplish, and removes the most reliable channel it has for checking its own
work. 
We therefore train a
\emph{hybrid} CUA that acts through two action channels: the GUI channel, with
visual observation and coordinate control, and the terminal channel, a
persistent shell environment in which it may run code, invoke
command-line utilities, and call application libraries directly.


Hybrid access does not by itself produce hybrid competence, even for
frontier teachers. Prior work shows that access to application tools does not
ensure their effective use, and that the same tool set can affect different
models differently \citep{osworldmcp,screenshotsortools}. The policy must learn
which channel is appropriate for the current operation and at what level of
abstraction to act. Effective hybrid execution routes each operation to the
channel where it fits best: coding and terminal execution for structured files
(such as updating a spreadsheet with one library call rather than dozens of
clicks), and GUI interaction for visual or live application state (such as
browser settings that a running browser caches in memory and overwrites on
exit). Because even frontier models exhibit suboptimal channel routing under a
naive setup, distilling their default rollouts passes those same limits on to
the student. The harness is not a neutral container: its tool interfaces,
context handling, and guidance directly shape how a fixed teacher model acts.
Evolving the harness before large-scale rollout elicits stronger hybrid
execution than the teacher achieves by default and carries those improvements
across the entire training corpus.

Moreover, whereas existing pipelines such as ProCUA and ScaleCUA~\citep{procua,scalecua}
post-train models that already have strong grounding, coding, and agent
capabilities, we present an end-to-end, progressive five-stage curriculum
starting from a general multimodally aligned checkpoint: GUI-oriented
midtraining, GUI agent SFT, coding and terminal SFT, hybrid CUA SFT, and
online hybrid RL. Sustaining such a
multi-stage curriculum requires massive, diverse training environments and
trajectories, for which we propose the \genname{} environment synthesis
pipeline. Unlike prior agent-first approaches that rely on open-ended agent
execution to construct each task from scratch, \genname{} uses an authoring
agent to construct application-specific preparers and reusable generation
workflows tailored to each application and its atomic functionalities, scaling
task and environment synthesis with high consistency and low recurring cost.

We introduce \sysname{}, a hybrid CUA trained on 318,837 teacher trajectories
over 293,806 synthesized tasks and environments that cover 6,915 atomic skills
in an agent-constructed skill taxonomy. Three design decisions define the
pipeline:



\begin{enumerate}[leftmargin=*,align=left]

\item \textbf{Learning from an evolved hybrid CUA harness.}
  To collect high-quality demonstrations even when a frontier teacher's default
  channel routing is imperfect, we evolve the harness including its scaffold, tool
  interfaces, prompts, and terminal execution environment, before large-scale
  rollout. Evolution here is as much pruning as growth: unlike skill libraries
  that continuously accumulate tools~\citep{voyager,skillweaver,eemcp}, we
  remove overlapping or ill-suited tools so the teacher produces clean hybrid
  trajectories that yield much larger gains when distilled into the student
  (\S\ref{sec:harness}, \S\ref{sec:harness-transfer-ablation}). Moreover,
  harness evolution produces specialized configurations with different tool
  sets---including a GUI harness, a terminal harness, and the hybrid CUA
  harness---allowing us to roll out the same synthesized tasks under multiple
  harnesses to collect trajectories emphasizing distinct capabilities across
  our training stages (\S\ref{sec:trainpipeline}).

\item \textbf{A single policy that learns the hybrid execution policy.}
  \sysname{} observes full-desktop screenshots and emits coordinate mouse and
  keyboard actions, and in the same episode may execute code and terminal
  commands on the machine it is operating. The two channels are complementary: pixels are the only
  channel that reaches spatial, canvas, and rendering-dependent state, while code
  edits at the abstraction the instruction is written in and lets the agent check
  its work against ground truth rather than against a screenshot of a dialog.
  Unlike systems that route between a GUI operator and a coding operator at the
  orchestrator level \citep{coact1,stateact}, the choice is internal to the policy
  and is therefore something the training data must teach (\S\ref{sec:agent}).

\item \textbf{Workflow-first task and environment synthesis.}
  Agent effort is spent once per \emph{skill} rather than once per task. For
  each atomic skill, an authoring agent constructs and debugs a reusable
  generation workflow that combines LLM nodes with application-specific
  \emph{preparers}, which materialize input artifacts such as a Calc workbook
  or a Thunderbird profile with a populated mail store. Combined workflows
  produce a task package of instruction, environment setup script, and rubric,
  and gate nodes check its consistency before the task is accepted. Shared
  workflows and preparers propagate each improvement across all subsequently
  generated task instances.
  Task synthesis, including skill-based generation, has also been studied
  in terminal settings, where the environment is a container
  filesystem \citep{terminalworld,skillsynth}; on a full desktop the environment
  is application state, which is what the preparer exists to produce
  (\S\ref{sec:synth}).

\end{enumerate}

With 4B parameters, \sysname{} achieves 66.0\% on OSWorld-Verified under a
50-step budget, outperforming larger open-source agents including Muse
Glimmer-30B (65.9\%), Qwen3.6-27B (63.9\%), and EvoCUA-32B (56.7\%).
Across our curriculum, performance climbs by 20.44 percentage points from GUI
SFT (45.56\%) through Hybrid SFT (55.84\%) to final RL (66.0\%), pushing
forward practical on-device intelligence for autonomous desktop workflows.

%% file: sections/2_related_work.tex
\section{Related Work}
\label{sec:related}

\paragraph{Computer-use agents and evaluation.}
GUI grounding methods such as SeeClick~\citep{seeclick} learn to locate
interface elements from language instructions. OSWorld~\citep{osworld}
and WebVoyager~\citep{webvoyager} evaluate the additional requirements of
multi-step interaction with desktop and browser environments. OpenCUA~\citep{opencua}
provides open training foundations for computer-use agents. Hybrid approaches
extend visual interaction with coding and terminal actions: CoAct-1~\citep{coact1}
uses an orchestrator to coordinate GUI and coding agents, whereas our student
selects GUI and terminal actions within a single policy. Tool availability alone
does not determine effective channel selection, as also examined in
OSWorld-MCP and controlled screenshot/tool comparisons~\citep{osworldmcp,screenshotsortools}.

\paragraph{Synthesized Environment.}
OS-Genesis~\citep{osgenesis} constructs GUI trajectories through reverse task
synthesis, while CUA-Gym~\citep{cuagym} co-generates task instructions,
environment states, and reward functions through an adversarial generator--discriminator
loop, and RecreationWorld~\citep{recreationworld} builds verifiable multi-platform
recreation environments for hybrid computer-use agents. \rev{For browser
environments, InfiniteWeb~\citep{zhang-etal-2026-infiniteweb} synthesizes
functional websites from specifications, and WebHarbor~\citep{webharbor2026}
reconstructs self-hosted website replicas for controllable evaluation and
training.} Rather than repeating an open-ended agent construction loop for
every task, \genname{} uses an authoring agent to build reusable
application-specific preparers and skill-conditioned workflows that construct
files, application state, \rev{and deterministic web replicas} at scale.

\paragraph{CUA Posttraining.}
ProCUA~\citep{procua} and ScaleCUA~\citep{scalecua} study scalable
computer-use post-training, but start from checkpoints that already possess
strong GUI grounding, coding, and agentic capabilities. MolmoWeb~\citep{molmoweb}
also documents an end-to-end training pipeline starting from a multimodally
aligned checkpoint, yet its scope is restricted to browser-only web agents and
lacks coverage of general desktop applications and hybrid GUI--terminal
execution. We instead present a progressive five-stage training curriculum
starting from a general multimodally aligned checkpoint across full desktop and
browser workflows.

%% file: sections/3_datagen_0921.tex
\section{Synthesized Data Generation for Hybrid Computer Use}
\label{sec:method}

\begin{figure}[t]
    \centering
    \includegraphics[width=\linewidth]{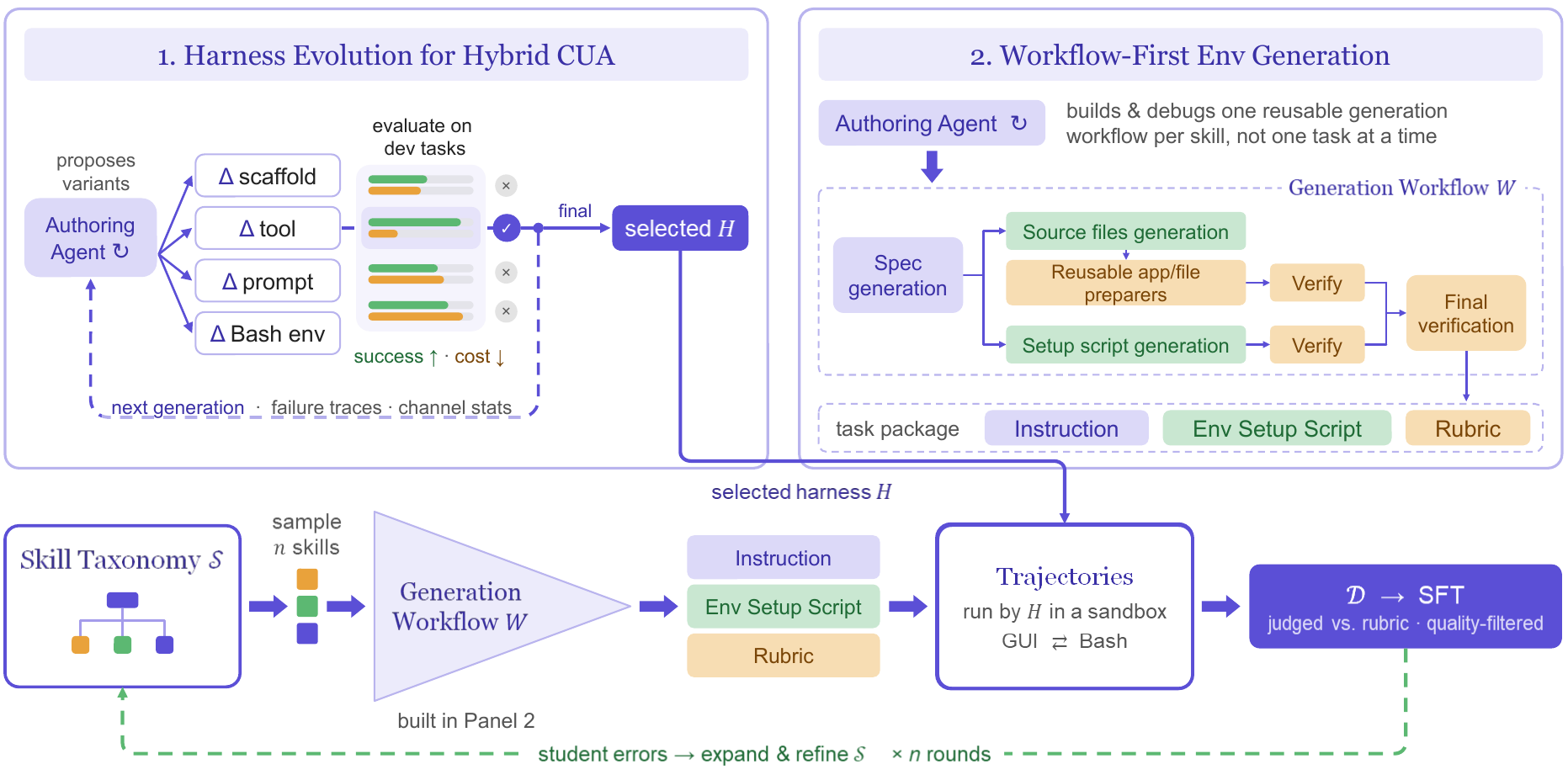}
    \caption{Schematic overview of the data generation pipeline. (1)~An authoring agent
    evolves the hybrid computer-use harness by proposing scaffold, tool-interface,
    prompt, and terminal-environment variants and selecting the harness $H$ with the best
    success--cost trade-off on development tasks. (2)~An authoring agent builds
    one reusable generation workflow $W$ per skill. A shared specification
    coordinates source-file and setup-script branches, whose preparers and gates
    produce a task package of instruction, environment setup script, and rubric.
    Bottom: skills sampled from the taxonomy $\mathcal{S}$ are instantiated
    through $W$, the teacher agent with $H$ executes the tasks in a sandbox
    through GUI and terminal actions, and rubric-judged, quality-filtered
    trajectories form $\mathcal{D}$ for SFT. Student performance feedback guides
    adaptive expansion and refinement of $\mathcal{S}$ across generation rounds.}
    \label{fig:method-overview}
\end{figure}

We construct supervised fine-tuning data through a pipeline that combines a
skill taxonomy $\mathcal{S}$, reusable generation workflows $W$, and an evolved
hybrid computer-use harness $H$. Each skill describes an atomic capability in a
specific application. We sample several skills to form a composite capability
and combine their workflows to generate a task instruction, an environment
setup script, and an evaluation rubric. Executing the setup script prepares a
sandbox with the preconditions needed to attempt the task. A teacher agent,
whose harness is optimized before large-scale trajectory collection, then
executes the task. Rubric-based LLM judging and trajectory-quality filtering
produce the retained trajectory dataset $\mathcal{D}$, which is used to train
the student model $M$. Student performance feedback guides adaptive expansion
and refinement of the skill taxonomy across generation rounds.
Figure~\ref{fig:method-overview} gives an overview of the pipeline. We first
describe teacher-agent evolution, then workflow-first generation, and finally
the taxonomy feedback loop.

\subsection{Evolution of the Hybrid Computer-Use Harness}
\label{sec:harness}
\label{sec:agent}

Collecting high-quality hybrid trajectories requires more than a strong
frontier model: the teacher must also select, sequence, and combine GUI actions
and terminal commands effectively, a strategy we call the \emph{hybrid execution
policy}. Under a standard setup, even frontier models show suboptimal channel
habits---such as clicking through repetitive file edits or modifying files on
disk that a running application caches in memory. Distilling directly from such
rollouts passes the teacher's default limitations on to the student.

We improve the teacher's hybrid execution by evolving its external harness while
keeping the underlying model weights fixed. We write the teacher agent as
\begin{equation}
    A_T = (M_T, H), \qquad H = (F, \mathcal{T}, P, \mathcal{E}),
    \label{eq:harness}
\end{equation}
where $M_T$ is the teacher model and the harness $H$ consists of four
components. The scaffold $F$ organizes the execution loop and context handling.
The tool interface $\mathcal{T}$ exposes the actions of each action
channel---GUI mouse and keyboard actions and terminal commands---together with
their argument schemas. The prompt $P$, i.e., the system prompt, guides tool
use and the hybrid execution policy. The terminal execution environment
$\mathcal{E}$ comprises the command-line utilities and application libraries
preinstalled in the Docker sandbox and accessible through the terminal channel.
Operational deployments require explicit authorization and sandbox isolation for
terminal execution.

Harness evolution optimizes $F$, $\mathcal{T}$, $P$, and $\mathcal{E}$ on a
shared development set. An authoring agent uses rubric-based task success,
per-task token cost, execution feedback, and application-level channel
statistics to select revisions that improve the success--cost trade-off.

Harness evolution includes removing redundant tools as well as adding useful
ones. Removing \texttt{xdotool} in favor of native screenshot-grounded GUI
actions avoids duplicate mouse/keyboard routes and fragile window-focus
handling that would split supervision across unequal choices. Conversely,
providing domain libraries such as \texttt{python-pptx} in the terminal
environment lets the teacher operate directly on slides, shapes, and text runs
rather than manually editing unpacked OOXML files.

\begin{figure}[!htbp]
    \centering
    \input{figures/harness_evolution_panel}
    \caption{Teacher--harness evolution across channel, scaffold, model, and
    prompt refinements. Left: task success and execution length for selected
    configurations. Right: the refinement dimension at each transition.
    Execution length is measured in model turns for $\alpha$ and $\beta_2$, and
    in tool calls for $\gamma_2$. Channel, scaffold, and prompt refinements
    modify $H$; model transitions update $M_T$.}
    \label{fig:harness-evolution}
\end{figure}
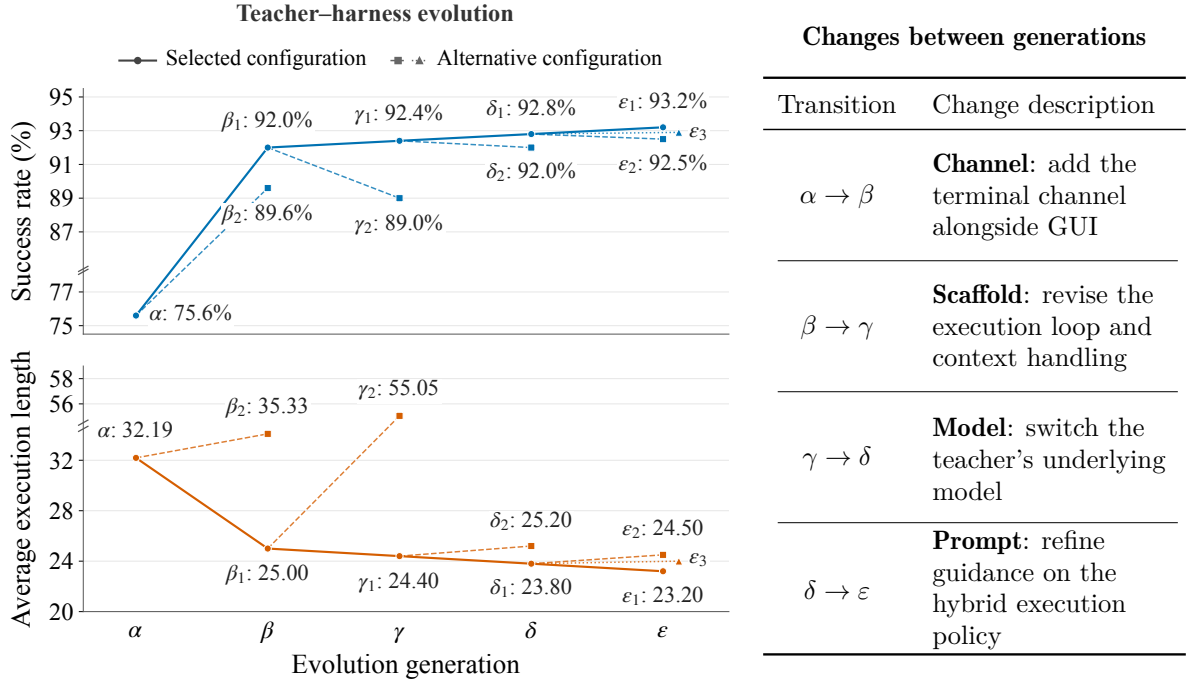

The GUI-and-coding variant achieves higher task success with fewer execution
steps than its AX-augmented counterpart; as shown in
Section~\ref{sec:harness-transfer-ablation}, this advantage widens into a
12.5-percentage-point gain when distilled into the 4B student. This comparison
guides joint optimization of the tool interface and prompt for effective hybrid
execution. Figure~\ref{fig:harness-evolution} presents selected
teacher--harness configurations across channel, scaffold, teacher-model, and
prompt refinements.

Once selected, the harness remains essentially fixed across subsequent
data-generation rounds. We use \texttt{gpt-5.6-terra} as the rubric judge to
verify task completion, while a trajectory-quality filter retains
demonstrations with clean channel routing and tool use. Optimizing the harness
before large-scale rollout ensures that $\mathcal{D}$ captures strong hybrid
execution for training $M$ (Section~\ref{sec:trainpipeline}).

\subsection{Workflow-First Task and Environment Generation}
\label{sec:synth}

A computer-use task package combines an instruction, an executable environment
setup, and an evaluation rubric. Workflow-first generation reuses the
construction procedure for these three coordinated components across task
instances.

\genname{} builds a reusable generation workflow for each atomic skill. Each
workflow combines LLM nodes, file preparers, and setup operations. For a
composite capability, the workflows share a unified specification of the
requested outcome, required files, and application preconditions, producing a
coordinated task package.

The source-file and setup-script branches share the task specification.
Branch-specific checks feed a final consistency gate over the complete task
package.

The workflow routes each required file to a preparer appropriate for its type,
reusing application-specific preparation logic across skill workflows. These
preparers materialize the input artifacts needed by the task. Other workflow
nodes produce the remaining setup code and combine it with shared setup
utilities, such as application launching and desktop-state adjustment. The
result is an \emph{environment setup script} (\emph{env setup script}), whose
purpose is to establish the required task preconditions and initial desktop
state inside the sandbox. Workflow-specific generation logic is thus separated
from reusable file and application preparation, as well as from generic setup
operations that are not tied to an individual skill.

Dedicated LLM nodes generate the user instruction and its evaluation rubric
from the shared specification. Task completion is judged by an LLM against this
rubric. Application setup details remain in the setup script, while the
instruction describes the user's intended outcome.

Validation operates at two levels. File-level gates check artifacts against the
shared specification, and a package-level gate checks consistency among the
instruction, setup requirements, and rubric. Gate feedback improves the
reusable workflow and shared preparers. Subsequent trajectory evaluation
assesses the teacher's execution of accepted tasks.

Workflow reuse separates one-time construction from recurring generation. LLM
nodes supply task-specific content within an established procedure, while
shared preparers provide reusable file and application setup logic across
skills and task instances.

\begin{revblock}
For browser and cross-application workflows, such as retrieving structured
records from a website and compiling them into a spreadsheet or document, live
websites introduce anti-bot blocks, content drift, and non-resettable backend
state. We therefore pair desktop application preparers with self-hosted,
deterministic web environments combining synthesized
InfiniteWeb~\citep{zhang-etal-2026-infiniteweb} sites and high-fidelity website
replicas built upon and extending WebHarbor~\citep{webharbor2026}, where each
replica resets from a local seed database per episode to support verifiable web
and cross-application task generation (Appendix~\ref{app:web-replicas}).
\end{revblock}

\subsection{Iterative Skill Taxonomy Expansion and Refinement}
\label{sec:taxonomy}

An exploration agent builds the initial skill taxonomy $\mathcal{S}$ by
traversing desktop applications and recording their atomic skills. Rather than
keeping $\mathcal{S}$ fixed, we refine and expand it across generation rounds
using feedback from student evaluations, adjusting skill granularity,
enriching descriptions, and adding application-specific variations where
needed.

The first generation round balances task coverage across atomic skills.
Composite capabilities instantiate these skills through their workflows;
teacher collection and filtering produce the supervision used to train $M$.

Student evaluation identifies capability areas that need denser coverage in the
next round. We add or refine the corresponding skill entries in $\mathcal{S}$,
update their workflows, and generate additional tasks and teacher trajectories.
Keeping the evolved harness fixed while expanding $\mathcal{S}$ across three
rounds cleanly separates improving \emph{how the teacher executes tasks} from
improving \emph{which tasks the pipeline supplies}.

\subsection{Dataset Statistics}
\label{sec:dataset-statistics}

\genname{} produces 293,806 distinct tasks 
covering 6,915 atomic skills. Single-application and multi-application tasks
account for 61.76\% and 38.24\%, respectively. Teacher execution and quality
filtering yield 318,837 retained trajectories.

\begin{figure}[t]
    \centering
    \includegraphics[width=\linewidth]{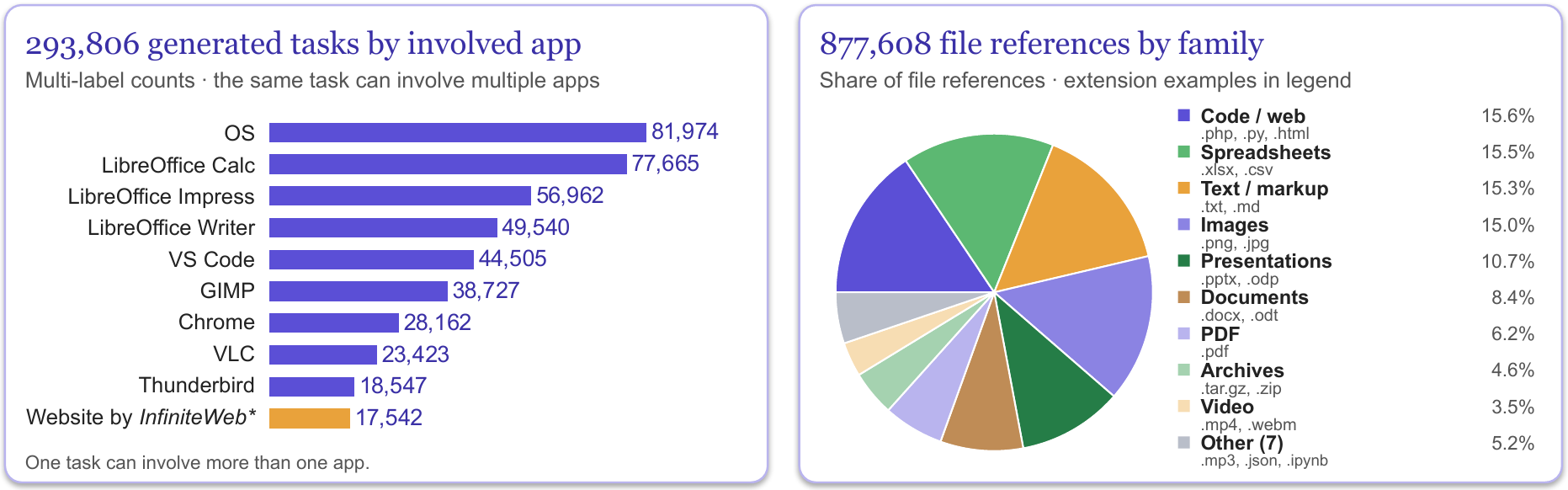}
    \caption{Application coverage and task--file reference families. Application counts
    measure multi-label application--task incidences. The 17,542
    InfiniteWeb~\citep{zhang-etal-2026-infiniteweb} tasks are reported separately. File-family shares
    are computed over 877,608 references.}
    \label{fig:dataset-coverage}
\end{figure}

The task inventory contains 877,608 task--file references, counted as distinct
extracted paths or filenames within each task. Selected reference families
include coding (136,736), spreadsheets (135,805), text or Markdown (134,050),
documents (74,052), PDFs (53,996), and presentations (93,784). File references occur in 266,030 tasks (90.5\%), averaging 2.99 references per
task. The inventory spans 278 suffix categories: 277 named suffixes and one
no-extension category. Figure~\ref{fig:dataset-coverage} summarizes application
involvement and reference families; Appendix~\ref{app:data-definitions} defines
the counting rules.

These synthesized tasks are rolled out under different harness
configurations, including the GUI harness, the terminal harness, and the
evolved hybrid CUA harness, to collect trajectory data that solve the tasks
through different channels and emphasize distinct capabilities for the
subsequent training stages described in Section~\ref{sec:trainpipeline}. Moreover, we also synthesize the reasoning content before tool call with a multi-stage annotation by GPT-5.4-mini. Within
the retained hybrid CUA dataset $\mathcal{D}$, supervision covers both execution
channels: GUI and terminal actions account for 68.19\% and 31.81\% of teacher
tool calls, respectively.

%% file: figures/harness_evolution_panel.tex
\begingroup
\noindent
\begin{minipage}[t]{0.635\linewidth}
    \vspace{0pt}
    \includegraphics[width=\linewidth]{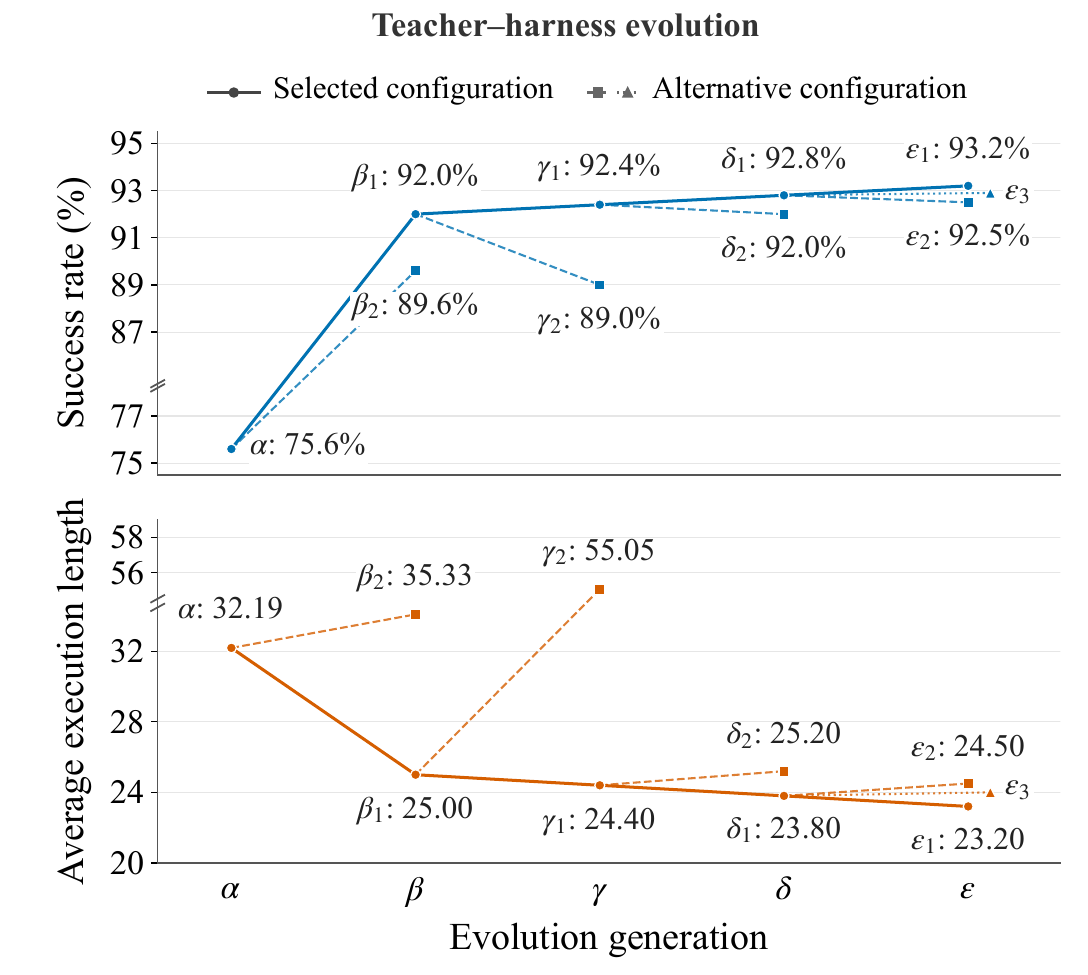}
\end{minipage}\hfill
\begin{minipage}[t]{0.345\linewidth}
    \vspace{12pt}
    \centering
    {\small\bfseries Changes between generations\par}
    \vspace{9pt}
    \small
    \setlength{\tabcolsep}{0pt}
    \renewcommand{\arraystretch}{1.15}
    \begin{tabular}{@{}>{\centering\arraybackslash}m{0.35\linewidth}@{\hspace{8pt}}>{\raggedright\arraybackslash}m{\dimexpr0.65\linewidth-8pt\relax}@{}}
        \toprule
        Transition & Change description \\
        \midrule
        \parbox[c][44pt][c]{\linewidth}{\centering $\alpha\rightarrow\beta$} & \parbox[c][44pt][c]{\linewidth}{\raggedright \textbf{Channel}: add the terminal channel alongside GUI} \\
        \cmidrule(lr{3pt}){1-2}
        \parbox[c][44pt][c]{\linewidth}{\centering $\beta\rightarrow\gamma$} & \parbox[c][44pt][c]{\linewidth}{\raggedright \textbf{Scaffold}: revise the execution loop and context handling} \\
        \cmidrule(lr{3pt}){1-2}
        \parbox[c][44pt][c]{\linewidth}{\centering $\gamma\rightarrow\delta$} & \parbox[c][44pt][c]{\linewidth}{\raggedright \textbf{Model}: switch the teacher's underlying model} \\
        \cmidrule(lr{3pt}){1-2}
        \parbox[c][44pt][c]{\linewidth}{\centering $\delta\rightarrow\varepsilon$} & \parbox[c][44pt][c]{\linewidth}{\raggedright \textbf{Prompt}: refine guidance on the hybrid execution policy} \\
        \bottomrule
    \end{tabular}
\end{minipage}
\endgroup

%% file: sections/4_trainpipeline.tex
\section{Training Pipeline of Mage-VL-Cowork}
\label{sec:trainpipeline}

Unlike pipelines that post-train checkpoints already specialized for GUI
grounding or coding, we present an end-to-end, progressive five-stage training
curriculum starting from an \rev{intermediate 4B \magevl{}} checkpoint:
(1)~GUI-Oriented Midtraining, (2)~GUI Agent SFT, (3)~Coding \& Terminal SFT,
(4)~Hybrid CUA SFT, and (5)~Online Hybrid RL.
This specialize-then-unify curriculum first builds solid, standalone
capabilities in visual grounding, GUI interaction, and coding and terminal
execution before teaching the model to route and combine both channels within a
single episode.

\paragraph{Context construction and training units.}
Context folding converts trajectories into turn-level examples, with
supervision applied to the next assistant response. Each context retains the
three most recent screenshots and four complete responses, while compressing
earlier steps into imperative \texttt{Action:} summaries immediately after the
system prompt. \rev{Each assistant response pairs a grounded \texttt{<think>}
rationale, a concise \texttt{Action:} summary, and the executed
\texttt{<tool\_call>}; together with the folded \texttt{Action:} history once
older screenshots slide out of the window, these structured reasoning, summary,
and tool-call traces act as a compact working memory, enabling the policy to
track prior intent and completed operations across long-horizon, interdependent
multi-step tasks (Appendix~\ref{app:context-relabeling}).}

Midtraining and SFT use two epochs over each dataset, packed into sequences
with a capacity of 32,768 tokens. Table~\ref{tab:training-compute} summarizes
the per-stage compute shares, trajectory counts, and token volumes;
Appendix~\ref{app:training-quantities} gives the data and compute-estimation
details.

\begin{table}[!htbp]
    \centering\small
    \begin{minipage}[c]{0.34\linewidth}
        \centering
        \includegraphics[width=\linewidth]{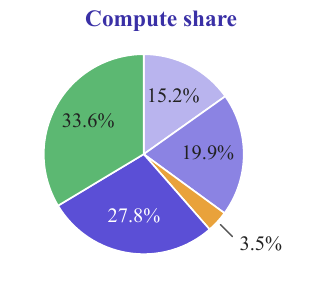}
    \end{minipage}\hfill
    \begin{minipage}[c]{0.64\linewidth}
        \centering
        \input{tables/training_compute}
    \end{minipage}
    \caption{Compute shares, training data, and nominal compute across the five
    stages. The pie chart (left) shows stage-wise compute shares, with colors
    corresponding to the stage rows. Tokens are reported in billions per epoch
    ($\mathrm{B}=10^9$ / epoch) of packed capacity, including padding and image
    tokens; midtraining and SFT use two epochs ($\mathrm{ZFLOPs}=10^{21}$
    FLOPs). Stage~1 packs multi-source examples and Stage~5 uses online RL
    rollouts, so their fixed trajectory or token entries are shown as dashes.
    Appendix~\ref{app:training-quantities} details the compute-estimation
    scope.}
    \label{tab:training-compute}
\end{table}

\subsection{Stage 1: GUI-Oriented Midtraining}
\label{sec:gui-midtraining}

\begin{revblock}
We initialize training from an intermediate 4B \magevl{} checkpoint. Evaluating
this starting checkpoint on the core atomic capabilities required for agentic
computer use---visual grounding, chain-of-thought reasoning, and structured tool
calling---reveals that all three remain underdeveloped: the model exhibits an
\emph{answer-machine} tendency~\citep{tong2024cambrian}, defaulting to terse
direct replies rather than reasoning before acting or emitting schema-compliant
tool calls. Stage~1 therefore introduces GUI-oriented midtraining to establish
these three atomic capabilities before multi-step trajectory post-training.

Targeting desktop and browser computer use, we curate an OS and web grounding
corpus from open-source GUI grounding
datasets including UI-E2I-Synth and others~\citep{gou2025uground,liu2025uie2isynth,wu2025osatlas,molmoweb,xie2025osworldg,seeclick},
formatting all spatial annotations as \texttt{computer\_use} \texttt{click}
tool calls and packing same-screenshot queries into shared sequences with
target-level deduplication~\citep{aguvis}. To jointly acquire screen
localization, structured function calling, and chain-of-thought reasoning
without mutual interference, we combine this grounding corpus with tool-calling
and GUI trajectory data during Stage~1.

To monitor whether midtraining checkpoints have acquired these atomic
capabilities without waiting for downstream post-training and multi-step
rollouts, we construct Single Step Cowork Bench directly from real downstream
desktop trajectories, using the target agent action space and system prompt to
provide fast diagnostic feedback. Appendix~\ref{app:midtraining-explorations}
details our midtraining data design and the two complementary recipes we
explored for reconciling non-reasoning grounding supervision with
reasoning-annotated tool calls, and Appendix~\ref{app:single-step-bench}
describes Single Step Cowork Bench.
\end{revblock}

\subsection{Stage 2: GUI Agent SFT}
\label{sec:gui-posttraining}

The second stage turns the visual and tool-format foundation into a GUI
execution policy. We collect teacher trajectories with the GUI-only harness. 
In this setting, task actions are
performed through the GUI channel, and the training examples pair the available
screenshot and interaction history with the teacher's next action. We use these
trajectories to fine-tune the midtrained checkpoint on complete computer-use
workflows.

GUI trajectories connect single-step grounding to multi-step interaction,
teaching the model which control to operate next, when to scroll, how to enter
text, and how to continue after the interface changes. Building strong
standalone GUI execution first ensures that when the model later learns hybrid
channel routing, it chooses between two capable channels rather than routing
around a weak visual policy.

\subsection{Stage 3: Coding \& Terminal SFT}
\label{sec:coding-enhancement}

The third stage develops coding and terminal execution capabilities, including
terminal operation, code generation, repository navigation, and
debugging with Terminal-Bench-style~\citep{pi2026dataengineeringscalingllm} and SWE-bench-style training data~\citep{openthoughts-agent}.

We combine general software-engineering data with desktop terminal harness
trajectories from productivity tasks we synthesized in Section~\ref{sec:method}. While repository-level tasks cover source
files, tests, and build tools, desktop coding trajectories teach the model to
read and modify workbooks, documents, presentations, and application state
using specialized Python libraries and command-line utilities.

Specifically, we run a terminal-only teacher harness on the same batch of
desktop tasks used for GUI trajectory collection in Stage~2. Retaining the
successful coding and terminal trajectories alongside general terminal and
software-engineering data gives the student direct examples of how to solve
target desktop tasks through code before Stage~4 combines both channels.

\subsection{Stage 4: Hybrid CUA SFT}
\label{sec:hybrid-training}

Stage~4 brings the visual and coding foundations together by training on
the 318,837 trajectories collected under the evolved hybrid harness
(Section~\ref{sec:harness}). Within a shared task context, a single policy
learns turn-level channel routing---deciding when to act through the GUI, when
to use coding and terminal execution, and when to switch between them, such as
using code to edit a structured file and then returning to the application
window to inspect the rendered result.

Because a 4B model is sensitive to noisy or ambiguous demonstrations,
optimizing the teacher harness for both task success and token efficiency and
filtering trajectories before training, which is critical. Rather than merely
exposing the student to a larger tool inventory, the resulting supervision
directly teaches effective channel routing and compact execution.

The final supervised stage draws on the 318,837 retained hybrid trajectories;
the final hybrid dataset contains 1,377,805 packed bins. The resulting model
combines the visual grounding and GUI execution learned in the first two stages,
the general and desktop-specific coding capabilities developed in the third,
and the hybrid execution policy demonstrated by the evolved hybrid agent.
We refer to the checkpoint obtained after this stage as Mage-VL-Cowork-SFT.

\subsection{Stage 5: Online Hybrid RL}
\label{sec:rl}

The final stage refines Mage-VL-Cowork-SFT with online reinforcement learning
in the hybrid harness. We train on the 1,988 tasks released by
ScaleCUA~\citep{scalecua}, each paired with a verifiable reward. Following
ScaleCUA, we use frontier sampling, with task weights based on pre-run success
rates to allocate rollouts toward the learning frontier. The policy is
optimized with GRPO~\citep{shao2024deepseekmath}. Rollouts are generated asynchronously with bounded
policy staleness, and we apply vanilla truncated importance sampling
(TIS)~\citep{yao2025tis} to correct the mismatch between the rollout and
training policies. Each rollout uses the same step budget as our OSWorld-Verified
evaluation. The resulting model is Mage-VL-Cowork, which we evaluate in
Section~\ref{sec:experiments}.

Each optimizer update admits 96 task groups with eight episode slots per group.
A four-batch admission window bounds policy lag at three updates, and TIS caps
importance weights at 2.0. Mixed-policy episodes are enabled. CPU workers
execute desktop environments, while GPU services handle training and inference.
The relay carries control, inference, and compact trajectories; weight
synchronization uses an intra-GPU-cluster path.
Section~\ref{sec:rl-infrastructure} summarizes the architecture;
Appendix~\ref{app:rl-system} gives the implementation details, runtime profile,
and measurement definitions.

%% file: tables/training_compute.tex
\definecolor{computeMid}{HTML}{B9B4EE}
\definecolor{computeGUI}{HTML}{8B83E3}
\definecolor{computeCode}{HTML}{E9A23B}
\definecolor{computeHybrid}{HTML}{5B4FD6}
\definecolor{computeRL}{HTML}{5CB872}
\newcommand{\computekey}[1]{\textcolor{#1}{\rule{0.65em}{0.65em}}\hspace{0.4em}}
\renewcommand{\arraystretch}{1.15}
\begin{tabular*}{\linewidth}{@{\extracolsep{\fill}}lrrr@{}}
    \toprule
    Stage & Trajectories & Tokens & \shortstack[r]{Compute\\(ZFLOPs)} \\
    \midrule
    \computekey{computeMid}1. GUI-Oriented Midtraining & \textemdash & 24.6B & 1.18 \\
    \computekey{computeGUI}2. GUI Agent SFT & 156.7k & 32.2B & 1.55 \\
    \computekey{computeCode}3. Coding \& Terminal SFT & 563.7k & 5.7B & 0.27 \\
    \computekey{computeHybrid}4. Hybrid CUA SFT & 318.8k & 45.1B & 2.17 \\
    \computekey{computeRL}5. Online Hybrid RL & \textemdash & \textemdash & 2.62 \\
    \midrule
    Total & \textemdash & \textemdash & 7.79 \\
    \bottomrule
\end{tabular*}

%% file: sections/4a_rl_infrastructure.tex
\section{Reinforcement Learning Infrastructure}
\label{sec:rl-infrastructure}

Our Rollout-as-a-Service (RaaS)~\citep{polar} infrastructure decouples desktop
execution, model inference, and policy optimization, integrating
PoLar~\citep{polar}, Slime, Azure Relay, and Azure Virtual Machine Scale Sets
(VMSS). CPU workers run environments and agent loops, an inference service
generates responses, and the training client consumes completed trajectories.
This separation supports asynchronous learning across long-running desktop
tasks; implementation details and runtime measurements are given in
Appendix~\ref{app:rl-system}.

\subsection{Workload Characteristics and Design Requirements}
\label{sec:infra-motivation}

Desktop rollouts combine inference with application loading, content entry, and
state updates. Individual observed episodes exceed 40 minutes, motivating
asynchronous scheduling so that completed tasks can advance training
independently of long-running episodes.

The resource requirements are also heterogeneous. Desktop environments need
CPU capacity, guest memory, and nested virtualization, whereas model inference
and optimization run on GPUs. For a compact model on B200 hardware,
environment execution and orchestration can remain important latency sources
even when individual model calls are relatively fast. We consequently manage
environment capacity separately from model-serving and training capacity,
and use asynchronous rollouts rather than requiring lockstep progress across
the desktop fleet.

Visual rollouts create repeated screenshot inputs across turn-level examples.
Compact episode transfer, prefetching, and node-local image expansion address
transport, preparation, and training-input materialization as separate stages
of the data path.

\subsection{System Architecture and Asynchronous Execution}
\label{sec:aaas-architecture}

\begin{figure}[H]
    \centering
    \includegraphics[width=\linewidth]{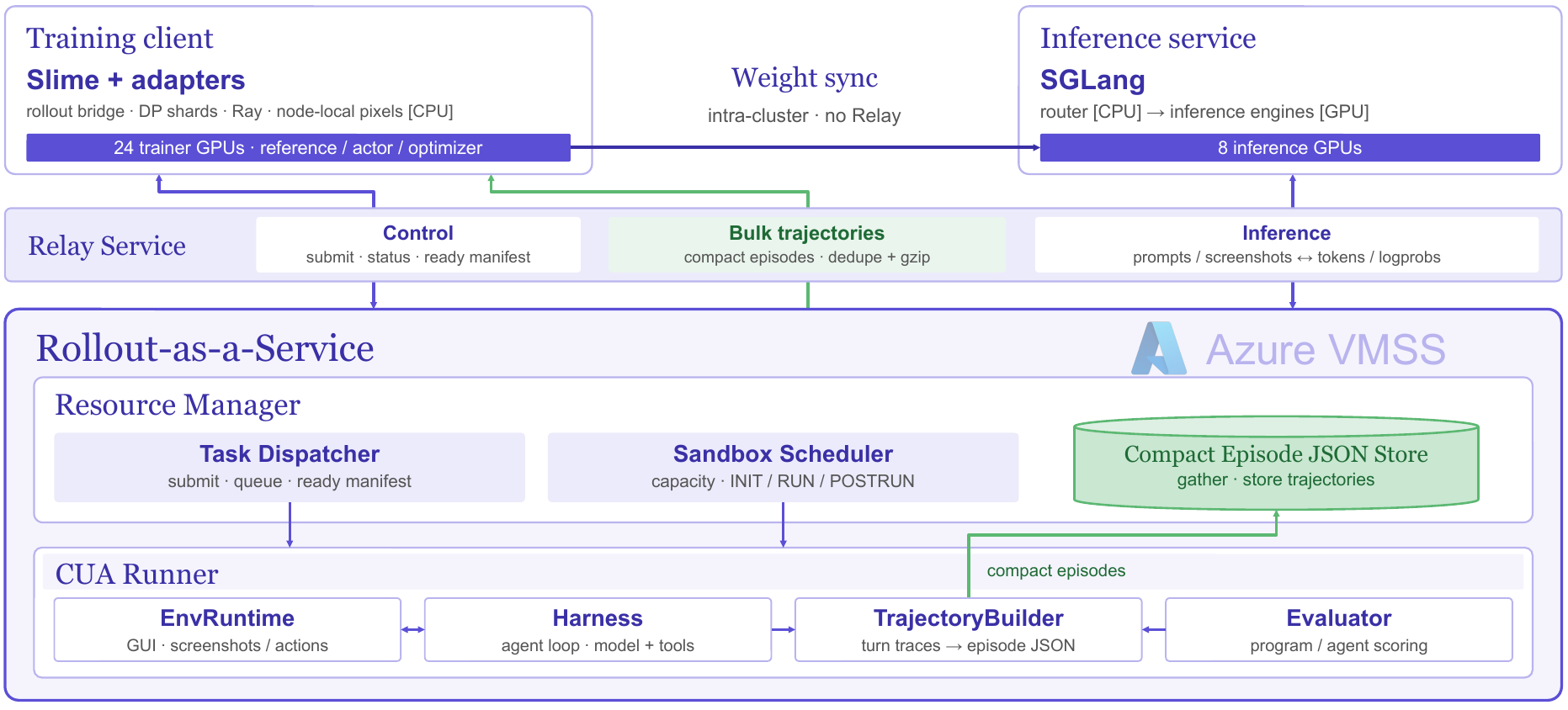}
    \caption{Rollout-as-a-Service architecture for asynchronous desktop RL. The
    resource manager coordinates task dispatch, sandbox scheduling, and compact
    episode storage. Runners handle environment interaction, harness execution,
    trajectory construction, and evaluation. The relay separates control,
    inference, and trajectory traffic; model weights synchronize within the GPU
    cluster. The diagram shows a 24-training/8-inference-GPU deployment.}
    \label{fig:rl-architecture}
\end{figure}

Figure~\ref{fig:rl-architecture} separates training, inference, and environment
execution. Slime and rollout adapters form the training client;
SGLang~\citep{sglang} serves model requests; and RaaS workers on Azure VMSS
host the resource manager and CUA runners. The resource manager dispatches
tasks, schedules sandboxes, and stores completed compact episodes. Each runner
coordinates environment interaction, the stateful harness, evaluation, and
trajectory construction. Control messages, inference traffic, and bulk
trajectory transfer use separate Azure Relay channels; weight synchronization
remains inside the GPU cluster.

Rollouts and optimization proceed asynchronously under the four-batch
admission window and three-update lag bound described in Section~\ref{sec:rl}.
Compact, deduplicated trajectories are transferred before per-turn sample
construction and node-local image expansion. Prefetching overlaps data
preparation with training, while lifecycle-specific admission limits manage
sandbox initialization, execution, and post-run processing independently.

The reference deployment provisions 1,120 GUI slots across 70 CPU hosts and
32 B200 GPUs, allocating 24 GPUs to training and eight to inference.
Appendix~\ref{app:rl-system} details deployment configurations and runtime
measurements.

%% file: sections/5_experiments.tex
\section{Experiments}
\label{sec:experiments}

Our experiments evaluate four properties of \sysname{}: desktop and browser
task performance at the 4B model scale, recurring task-generation expenditure,
performance--efficiency gains across training stages, and the relationship
between teacher-harness quality and student outcomes.

\subsection{Benchmarks and Evaluation Setup}
\label{sec:eval-benchmarks}

We evaluate on held-out task sets.

\textbf{WebVoyager-Static.}
\rev{Evaluating browser agents on live websites~\citep{webvoyager} is confounded by
both environment instability and unreliable grading: page content varies across
time and location, anti-bot checks and access blocks interrupt rollouts, VLM
judges lack backend ground truth and often credit unverified claims absent from
the final screenshot, and newer models already know much of the target
information and tend to guess answers without browsing. To construct a
high-fidelity offline benchmark, we build on WebHarbor~\citep{webharbor2026}
and substantially expand its self-hosted replicas, reconstructing the missing
pages, interactive controls, and local database records required to faithfully
support all 643 tasks across the 15 WebVoyager~\citep{webvoyager} sites. Rather
than relying on a final-screenshot VLM judge alone, grading verifies both the
execution trace, by confirming that the agent visited the target pages and
performed the required UI operations, and the final outcome, by checking the
agent's answer and backend database state against ground truth. This
deterministic setup eliminates the score inflation common on live web
benchmarks (Table~\ref{tab:webvoyager-static}) while running at a fraction of
the cost of full desktop VM rollouts, making WebVoyager-Static an effective
multi-step diagnostic for validating our web data and iterating on early
post-training checkpoints.}

\textbf{OSWorld-Verified.}
OSWorld-Verified evaluates end-to-end desktop task completion across documents,
spreadsheets, presentations, browsers, image editing, and operating-system
operations. We evaluate the final RL checkpoint on 361 held-out tasks with a
50-step budget, excluding Google Drive. The reported result averages two
evaluation runs. For each run, the overall score is
$100N^{-1}\sum_{i=1}^{N}s_i$, where $N=361$ and $s_i$ is the benchmark score
awarded to task $i$. Table~\ref{tab:osworld-main-results} reports the overall
scores and available category breakdowns.

\subsection{Main Results on OSWorld-Verified}
\label{sec:osworld-main-results}

\sysname{} achieves 66.0\% on OSWorld-Verified with a 4B policy and a 50-step
budget. Table~\ref{tab:osworld-main-results} places this result alongside
models across a broad range of parameter scales: our 4B policy outperforms
open-source agents up to 7.5--18$\times$ larger, including Muse Glimmer-30B
(65.9\%), Qwen3.6-27B (63.9\%), EvoCUA-32B (56.7\%), and OpenCUA-72B (44.9\%).
At the 4B scale, \sysname{} exceeds Fara1.5-4B (13.9\%) by 52.1 percentage
points overall, while achieving 87.5\% on OS and 62.7\% on multi-application
Workflow tasks, which is the highest OS score in the table and 14.6 percentage points
above ScaleCUA-Qwen3.5-9B (48.1\%) on Workflow. This capability-to-size
efficiency is the central outcome of our evolved-harness supervision and
progressive five-stage training curriculum. Baseline step budgets and all
reported scores are listed in the table.

\begin{table}[!htbp]
    \centering
    \small
    \resizebox{\linewidth}{!}{\input{tables/osworld_results.tex}}
    \caption{OSWorld-Verified results across model scales. \sysname{} achieves
    66.0\% with 4B parameters under a 50-step budget. Baseline scores and step
    budgets are shown as reported. $^\dagger$Evaluated by us in one run on
    361 tasks, excluding Google Drive. Blue shading marks the two highest
    overall scores in each model group.}
    \label{tab:osworld-main-results}
\end{table}

\input{sections/5a_cost_and_progression}
\input{sections/5b_execution_diagnostics}

\subsection{Browser Task Performance}
\label{sec:intermediate-evaluation}

\begin{revblock}
On WebVoyager-Static, the 4B \sysname{} SFT model achieves 58.6\% success
(377/643 tasks; Table~\ref{tab:webvoyager-static}), ranking second overall
behind the specialized browser agent Fara1.5-4B (68.0\%). Despite being trained
as a general desktop-and-browser agent, \sysname{} outperforms both larger
8B--9B models---such as Qwen3-VL-8B-Thinking~\citep{qwen3vl} (53.7\%) and
Qwen3.5-9B (53.3\%)---and dedicated open-source browser agents including
OpenWebRL-4B~\citep{openwebrl} (47.1\%), MolmoWeb-8B~\citep{molmoweb} (46.8\%),
and Fara-7B~\citep{fara7b} (40.4\%).

Comparing previously published live WebVoyager scores against our offline
evaluation reveals a drop of 12.8 to 33.1 percentage points across the five
models with self-reported live results~\citep{fara7b,openwebrl,molmoweb}.
Auditing the rollout traces under both the original live-judge protocol and our
offline verifier shows why: because the standard live judge inspects only the
final screenshot and explicitly instructs the evaluator to trust unsupported
claims in the agent's text (\emph{``choose to believe the content''}), it
rewards fluent guessing and masks three recurring failure modes in current web
agents: (1)~\emph{shortcut answering from search snippets or parametric
memory}, where agents answer after a few steps from a result card or prior
knowledge without opening the detail page that holds the ground-truth evidence;
(2)~\emph{incomplete multi-constraint filtering}, where agents locate the target
site and issue a keyword query but fail to operate interactive filter controls
(such as date pickers, price sliders, or sort menus), returning the first
plausible candidate that violates secondary constraints; and
(3)~\emph{uncommitted stateful interactions}, where agents fill out forms or
click preliminary buttons and declare success while the backend database records
no actual mutation (e.g., an empty cart or unsubmitted booking). These findings
suggest that screenshot-only judging can inadvertently incentivize premature,
persuasive answers over thorough interaction. By jointly verifying page-visit
traces, ground-truth constraints, and backend database state, WebVoyager-Static
removes these false positives and provides a more diagnostic reference for
evaluating genuine multi-step web interaction.
\end{revblock}

\begin{table}[!htbp]
    \centering
    \small
    \input{tables/webvoyager_static_results.tex}
    \caption{WebVoyager-Static results across all 643 tasks under identical
    offline evaluation settings\rev{, alongside self-reported live
    WebVoyager scores where available~\citep{fara7b,openwebrl,molmoweb}}. Blue
    shading marks the best and second-best WebVoyager-Static success rates.}
    \label{tab:webvoyager-static}
\end{table}

\FloatBarrier
\subsection{Effect of the Evolved Harness on Student Performance}
\label{sec:harness-transfer-ablation}

We next evaluate how evolving the harness around a fixed frontier teacher model
shapes both teacher execution and downstream 4B student learning. Starting from
the official GPT-5.4 OSWorld agent, $\beta_1$ pairs GUI interaction with a
coding channel, whereas $\beta_2$ exposes GUI, coding, and accessibility (AX)
channels. Both teachers use the same underlying GPT-5.4 model, and each
generates the trajectories used to train its corresponding student. All four
agents are evaluated on the same 672 held-out multi-application tasks under
identical evaluation settings (Table~\ref{tab:harness-transfer}).

\begin{table}[!htbp]
    \centering\small
    \input{tables/harness_transfer}
    \caption{Teacher--student comparison for two hybrid harness variants.
    All four agents are evaluated on the same 672 held-out multi-application
    tasks. Each student is trained on trajectories generated by the corresponding
    teacher variant. Higher success rates and fewer steps are preferred;
    bold marks the better student result for each metric.}
    \label{tab:harness-transfer}
\end{table}

With the teacher model held fixed, the $\beta_1$ harness outperforms the
AX-augmented $\beta_2$ harness on both success rate (92.0\% vs.\ 89.6\%,
+2.4\,pp) and step count (25.0 vs.\ 35.3 steps, $-29.2\%$). This gap widens
substantially when distilled into the 4B students: the $\beta_1$ student
reaches 37.6\% versus 25.1\% for $\beta_2$, while using 21.5\% fewer steps (30.91 vs.\ 39.4). While a frontier
teacher can partially absorb an extra channel by taking more steps, a compact
student is much more sensitive to ambiguous channel routing. Refining and
pruning the teacher harness before rollout therefore yields student gains far
larger than the teacher's raw success-rate difference alone would suggest.

\FloatBarrier

%% file: tables/osworld_results.tex
\begingroup
\definecolor{cwBest}{HTML}{739EC4}
\definecolor{cwSecond}{HTML}{94B6D3}
\setlength{\tabcolsep}{6pt}
\renewcommand{\arraystretch}{1.10}
\begin{tabular}{@{}lcrrrrrr@{}}
\toprule
Model & Steps & OS & Office & Daily & Professional & Workflow & Overall \\
\midrule
Human & \textemdash & \textemdash & \textemdash & \textemdash & \textemdash & \textemdash & 72.4 \\
\multicolumn{8}{@{}l}{\textit{Frontier models}} \\
GPT-5.5 & \textemdash & \textemdash & \textemdash & \textemdash & \textemdash & \textemdash & \cellcolor{cwBest}\textbf{78.7} \\
Claude Opus 4.7 & \textemdash & \textemdash & \textemdash & \textemdash & \textemdash & \textemdash & \cellcolor{cwSecond}\underline{78.0} \\
GPT-5.4 & \textemdash & \textemdash & \textemdash & \textemdash & \textemdash & \textemdash & 75.0 \\
Kimi-K2.6 & \textemdash & \textemdash & \textemdash & \textemdash & \textemdash & \textemdash & 73.1 \\
GPT-5.4-mini$^\dagger$ & 50 & 79.2 & 68.0 & 76.6 & 69.4 & 63.3 & 69.6 \\
Claude Sonnet 4.5 & 100 & 70.8 & 72.6 & 61.4 & 63.2 & 49.5 & 62.9 \\
Seed-1.8 & 100 & 66.7 & 68.8 & 67.1 & 71.4 & 42.4 & 61.9 \\
Agent S2.5 w/ o3 & 50 & 75.0 & 53.8 & 54.0 & 75.5 & 39.5 & 54.2 \\
UI-TARS-2 & 100 & 41.7 & 61.1 & 62.1 & 61.2 & 34.1 & 53.1 \\
Claude Sonnet 4 & 50 & 45.8 & 43.5 & 55.1 & 55.1 & 28.5 & 43.9 \\
OpenAI CUA & 50 & 70.8 & 23.0 & 37.2 & 51.0 & 15.8 & 31.3 \\
\midrule
\multicolumn{8}{@{}l}{\textit{Models with reported parameter counts}} \\
ScaleCUA-Qwen3.5-9B & 50 & 79.2 & 70.6 & 74.3 & 89.5 & 48.1 & \cellcolor{cwBest}\textbf{68.7} \\
Muse Glimmer-30B & \textemdash & \textemdash & \textemdash & \textemdash & \textemdash & \textemdash & 65.9 \\
Qwen3.6-27B & \textemdash & \textemdash & \textemdash & \textemdash & \textemdash & \textemdash & 63.9 \\
EvoCUA-32B & \textemdash & \textemdash & \textemdash & \textemdash & \textemdash & \textemdash & 56.7 \\
Qwen3.5-27B & \textemdash & \textemdash & \textemdash & \textemdash & \textemdash & \textemdash & 56.2 \\
ComputerRL-9B & 50 & 75.0 & 45.8 & 52.5 & 71.4 & 27.9 & 48.0 \\
ProCUA-SFT & \textemdash & \textemdash & \textemdash & \textemdash & \textemdash & \textemdash & 45.0 \\
OpenCUA-72B & 50 & 45.8 & 42.7 & 53.8 & 75.5 & 22.3 & 44.9 \\
DART-GUI-7B & 30 & 62.5 & 39.3 & 50.8 & 73.5 & 16.7 & 42.1 \\
Qwen3.5-9B & \textemdash & \textemdash & \textemdash & \textemdash & \textemdash & \textemdash & 41.8 \\
OpenCUA-32B & 50 & 60.9 & 29.9 & 41.0 & 65.3 & 14.6 & 35.1 \\
Fara1.5-4B & 50 & 25.0 & 7.6 & 29.5 & 12.2 & 6.6 & 13.9 \\
\midrule
\textit{Mage-VL-Cowork-4B} & 50 & 87.5 & 64.9 & 65.0 & 66.3 & 62.7 & \cellcolor{cwSecond}\underline{66.0} \\
\bottomrule
\end{tabular}
\endgroup

%% file: sections/5a_cost_and_progression.tex
\subsection{Task-Generation Expenditure}
\label{sec:generation-cost}

Workflow-first generation reduces recurring per-task expenditure by 82.6\%,
from \$0.906 to \$0.158, and token consumption by 93.3\%, from 1,276K to 85K
tokens (Table~\ref{tab:generation-cost}). Both approaches use
\texttt{gpt-5.6-terra}. Reusing generation workflows and application preparers
shifts repeated construction work into reusable artifacts.

\begin{table}[htbp]
    \centering\small
    \begin{tabular}{@{}lrr@{}}
        \toprule
        Env Synth. Method & Tokens per task & Expenditure per task (USD) \\
        \midrule
        Agent-first & 1,276K & 0.906 \\
        \genname{} & 85K & 0.158 \\
        \bottomrule
    \end{tabular}
    \caption{Recurring task-generation expenditure per task. \genname{}
    reuses workflows and application preparers; one-time workflow
    authoring and debugging are excluded from the per-task comparison.}
    \label{tab:generation-cost}
\end{table}

\FloatBarrier
\subsection{Training Progression}
\label{sec:training-progression}

\begin{figure}[htbp]
    \centering
    \includegraphics[width=\linewidth]{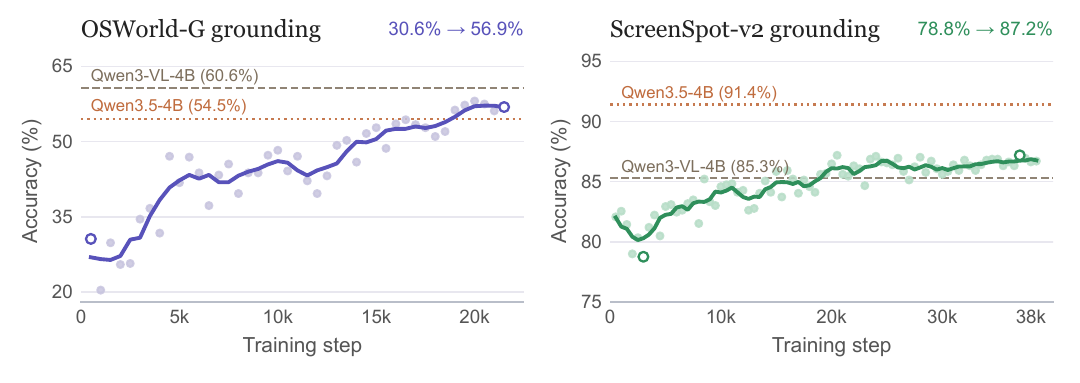}
    \caption{\rev{Grounding progression during Stage~1 GUI-oriented midtraining
    on OSWorld-G (left) and ScreenSpot-v2 (right). Dots show individual
    checkpoint evaluations, solid curves show five-checkpoint moving averages,
    and horizontal lines mark the Qwen3-VL-4B and Qwen3.5-4B reference
    baselines.}}
    \label{fig:midtraining-grounding}
\end{figure}

\rev{During Stage~1 GUI-oriented midtraining, screen grounding accuracy rises
steadily as the policy acquires visual localization prior to multi-step
trajectory training (Figure~\ref{fig:midtraining-grounding}): from 30.6\% to
56.9\% on OSWorld-G~\citep{xie2025osworldg}, surpassing Qwen3.5-4B (54.5\%), and
from 78.8\% to 87.2\% on ScreenSpot-v2~\citep{wu2025osatlas}, exceeding
Qwen3-VL-4B (85.3\%). Design trade-offs for Stage~1 midtraining and single-step
diagnostics are detailed in Appendices~\ref{app:midtraining-explorations}
and~\ref{app:single-step-bench}.}

\begin{figure}[htbp]
    \centering
    \includegraphics[width=\linewidth]{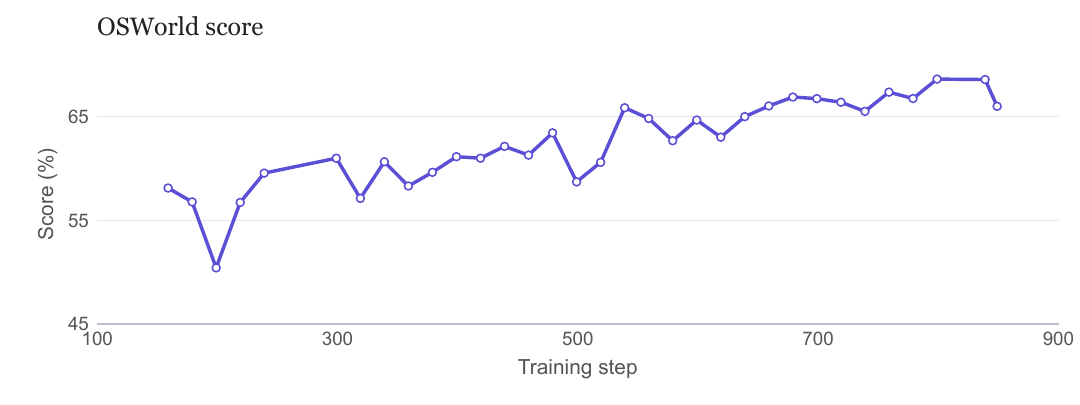}
    \caption{OSWorld-Verified scores during reinforcement learning, using
    all 33 recorded evaluations. The final model scores 66.0\%.}
    \label{fig:rl-checkpoints}
\end{figure}

Building on this midtrained checkpoint, end-to-end OSWorld-Verified performance
rises by 20.44 percentage points across the downstream training stages: from
45.56\% after GUI SFT to 55.84\% after Hybrid SFT (+10.28\,pp), and to 66.0\%
after online RL (+10.16\,pp). Figure~\ref{fig:rl-checkpoints} shows the
progression during RL, and Section~\ref{sec:execution-diagnostics} examines the
accompanying changes in execution length and token usage.

%% file: sections/5b_execution_diagnostics.tex
\FloatBarrier
\subsection{Execution Efficiency and Token Usage}
\label{sec:execution-diagnostics}

GUI SFT, Hybrid SFT, and Hybrid RL denote the final checkpoints of the GUI Agent SFT
stage, Hybrid CUA SFT stage, and RL stage, respectively. We compare task
performance, execution length, and token usage on 361 OSWorld-Verified tasks.

\begin{figure}[!htbp]
    \centering
    \captionsetup[subfigure]{format=plain,font=small,labelformat=parens,
        labelsep=space,justification=centering,singlelinecheck=false,skip=3pt}
    \begin{subfigure}[t]{0.291\linewidth}
        \centering
        \includegraphics[width=\linewidth]{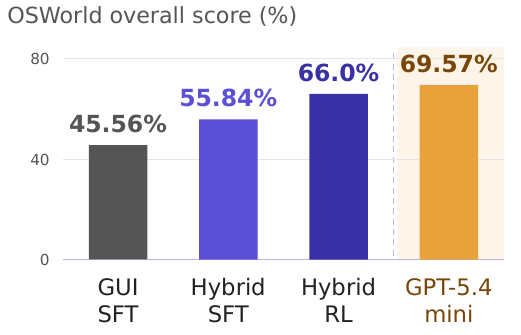}
        \caption{Task performance}
        \label{fig:historical-performance}
    \end{subfigure}\hfill
    \begin{subfigure}[t]{0.278\linewidth}
        \centering
        \includegraphics[width=\linewidth]{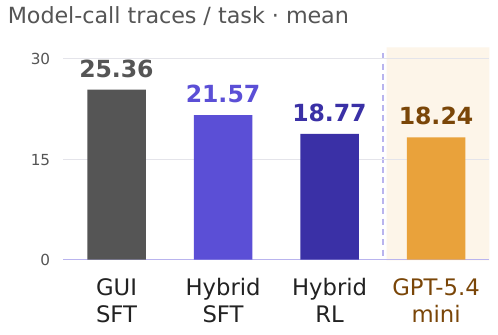}
        \caption{Execution length}
        \label{fig:historical-execution}
    \end{subfigure}\hfill
    \begin{subfigure}[t]{0.391\linewidth}
        \centering
        \includegraphics[width=\linewidth]{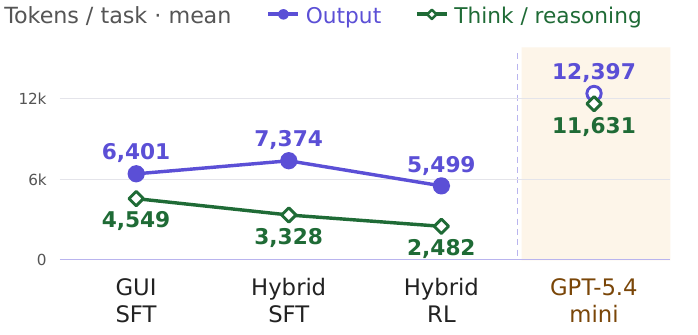}
        \caption{Token usage}
        \label{fig:historical-tokens}
    \end{subfigure}
    \caption{Stage-wise task performance, execution length, and token usage.
    Execution length counts model-call traces per task. Think/reasoning
    tokens are a subset of output tokens. Detailed mean, median, and P90
distributions are reported in Appendix~\ref{app:training-quantities}
(Table~\ref{tab:historical-distributions}).}
    \label{fig:historical-diagnostics}
\end{figure}
\FloatBarrier

Figure~\ref{fig:historical-diagnostics} shows that across the three training
stages, task scores rise from 45.56\% after GUI SFT to 55.84\% after Hybrid SFT
and 66.0\% after RL, while mean model calls decrease from 25.36 to 21.57 and
18.77 (GPT-5.4-mini averages 18.24). Hybrid SFT pairs a 10.28-percentage-point
gain and 14.9\% fewer model calls with higher mean output-token usage,
averaging 7,374 tokens per task compared with 6,401 for GUI SFT. RL then
improves both performance and generation efficiency: it adds 10.16 percentage
points over Hybrid SFT, reduces model calls by 13.0\%, and reduces both output
and think tokens by approximately 25.4\%, bringing mean output tokens to 5,499
per task. The final policy therefore combines stronger task performance with
shorter interaction sequences and lower output volume.

\FloatBarrier

%% file: tables/webvoyager_static_results.tex
\begingroup
\definecolor{cwBest}{HTML}{739EC4}
\definecolor{cwSecond}{HTML}{94B6D3}
\setlength{\tabcolsep}{6pt}
\renewcommand{\arraystretch}{1.10}
\begin{tabular}{@{}lrrr@{}}
\toprule
Model & WebVoyager-Static (\%) & \rev{Live reported (\%)} & \rev{$\Delta$ (pp)} \\
\midrule
\multicolumn{4}{@{}l}{\textit{Open-source models}} \\
Fara1.5-4B & \cellcolor{cwBest}\textbf{\rev{68.0}} & \rev{80.8} & \rev{$-12.8$} \\
Qwen3-VL-8B-Thinking~\citep{qwen3vl} & \rev{53.7} & \textemdash & \textemdash \\
Qwen3.5-9B & \rev{53.3} & \textemdash & \textemdash \\
Qwen3-VL-4B-Thinking~\citep{qwen3vl} & \rev{49.6} & \textemdash & \textemdash \\
OpenWebRL-4B~\citep{openwebrl} & \rev{47.1} & \rev{74.1} & \rev{$-27.0$} \\
MolmoWeb-8B~\citep{molmoweb} & \rev{46.8} & \rev{78.2} & \rev{$-31.4$} \\
MolmoWeb-4B~\citep{molmoweb} & \rev{42.6} & \rev{75.2} & \rev{$-32.6$} \\
Fara-7B~\citep{fara7b} & \rev{40.4} & \rev{73.5} & \rev{$-33.1$} \\
Qwen3.5-4B & \rev{37.3} & \textemdash & \textemdash \\
\midrule
\textit{Mage-VL-Cowork-SFT} & \cellcolor{cwSecond}\underline{\rev{58.6}} & \textemdash & \textemdash \\
\bottomrule
\end{tabular}
\endgroup

%% file: tables/harness_transfer.tex
\begin{tabular*}{\linewidth}{@{\extracolsep{\fill}}lrrrr@{}}
    \toprule
    & \multicolumn{2}{c}{Teacher} & \multicolumn{2}{c}{Student} \\
    \cmidrule(lr){2-3}\cmidrule(lr){4-5}
    Metric & $\beta_1$ & $\beta_2$ & $\beta_1$ & $\beta_2$ \\
    \midrule
    Success rate (\%) $\uparrow$ & 92.0 & 89.6 & \textbf{37.6} & 25.1 \\
    Steps $\downarrow$ & 25.0 & 35.3 & \textbf{30.91} & 39.4 \\
    \bottomrule
\end{tabular*}

%% file: sections/6_conclusion.tex
\section{Conclusion}
\label{sec:conclusion}

\sysname{} brings effective hybrid computer use to a 4B policy, achieving
66.0\% on OSWorld-Verified under a 50-step budget and outperforming
open-source agents several times its size. Rather than post-training a
checkpoint already specialized for computer use, our progressive five-stage
training curriculum, GUI-oriented midtraining, GUI agent SFT, coding and
terminal SFT, hybrid CUA SFT, and online hybrid RL, builds visual grounding,
GUI execution, desktop coding and terminal operation, hybrid channel routing,
and online RL from a general multimodally aligned starting point. Upstream of training, evolving the teacher harness, elicits stronger
demonstrations than the teacher model's default setup, while reusable
skill-conditioned workflows cut recurring per-task generation expenditure by
82.6\%. Together, these results provide a scalable route from reusable
environment construction and teacher-harness optimization to compact hybrid
computer-use policies, pushing forward the frontier of on-device intelligence.

%% file: sections/7-author.tex
\section*{Contributor List}
\label{sec:contributor}

\noindent\textbf{Contributors:} 
Xiaoyi Zhang$^{*\dagger}$, Haoqing Wang$^{*}$, Xinjie Zhang, Xiang Ming, Xing Wang, Senqiao Yang, Kaichen Zhang, Zhaoyang Jia, Andong Tan, Xinjiang Wang, Jinghao Guo, Yifei Shen, Xiao Li, Peng Zhang, Xun Guo, Jiahao Li, Shicheng Zheng, Jinglu Wang, Zongyu Guo, Wenxuan Xie, Bin Li, Yan Lu

\vspace{0.5em}
\noindent{\footnotesize $^{*}$Equal Contribution. \quad $^{\dagger}$Project Lead (xiaoyizhang@microsoft.com).}

%% file: sections/appendix.tex
\clearpage
\appendix
\input{sections/appendix_additions}

%% file: sections/appendix_additions.tex
\section{Web Replica Environments and Data Counting Definitions}
\label{app:data-definitions}

\subsection{High-Fidelity Web Replicas}
\label{app:web-replicas}

\begin{figure}[!htbp]
    \centering
    \includegraphics[width=\linewidth]{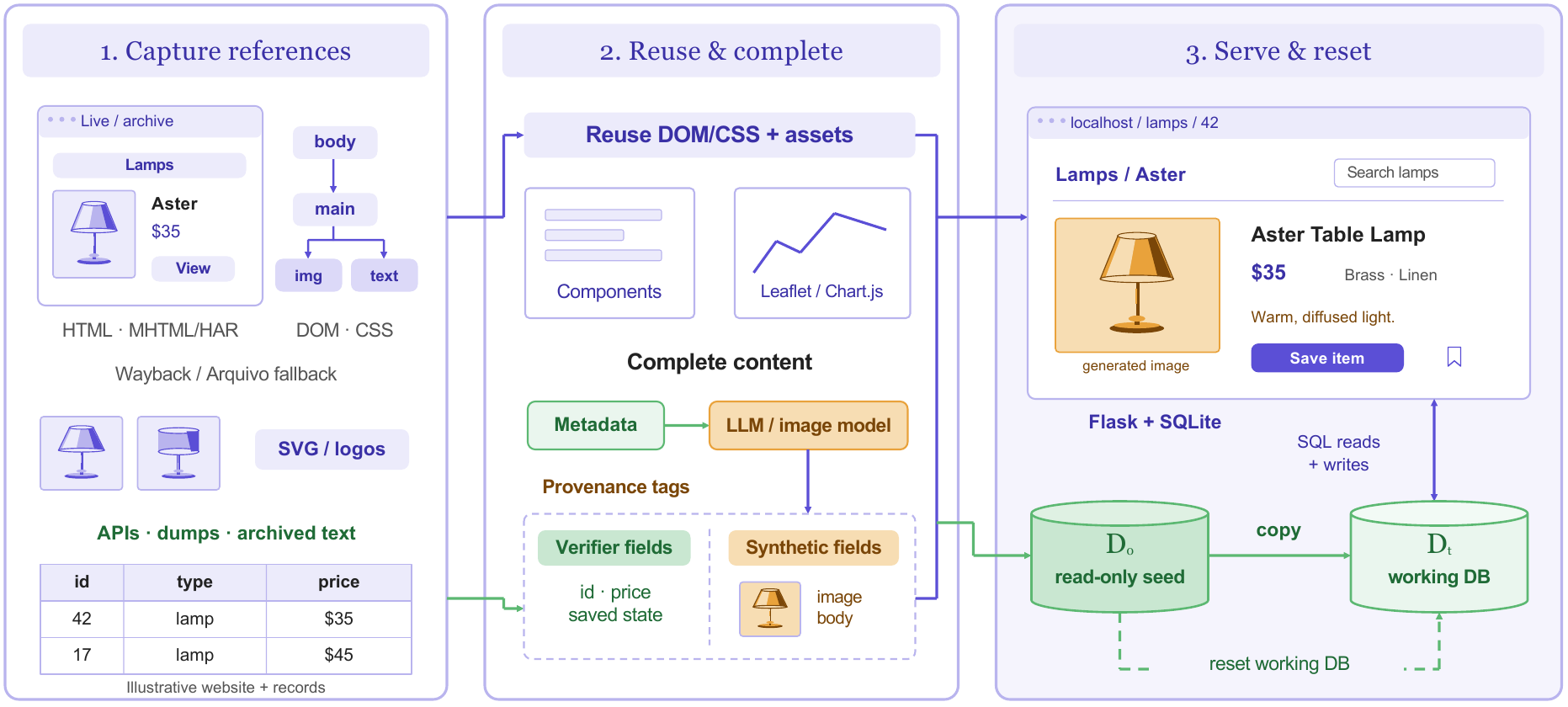}
    \caption{\rev{Construction and reset of a web replica. Reference structure,
    assets, and records are combined with generated content to assemble the
    local site. Each episode operates on a working database restored from the
    read-only seed. The website and records are illustrative.}}
    \label{fig:web-replica-pipeline}
\end{figure}

\begin{revblock}
To support the controllable browser workflows and offline evaluations introduced
in Sections~\ref{sec:synth} and~\ref{sec:eval-benchmarks}, we pair
synthesized InfiniteWeb~\citep{zhang-etal-2026-infiniteweb} environments with
self-hosted website replicas built upon and expanding
WebHarbor~\citep{webharbor2026} (Figure~\ref{fig:web-replica-pipeline}). Each
replica runs as a local \texttt{Flask} and \texttt{SQLite} service with target
domains routed to the local host. Before every episode, the environment restores
a fresh working database from a read-only seed snapshot, isolating stateful
operations across rollouts while exposing the underlying tables for deterministic
verification. Constructing and upgrading these replicas follows two core
principles:

\paragraph{Frontend and interactive fidelity.}
Early mock environments often simplify pages into generic templates or static
placeholder blocks, which reduces visual complexity and omits the controls
needed for multi-step tasks. To match real-world interfaces, we capture
reference snapshots from target websites---spanning DOM hierarchies, computed
CSS rules, and static assets (logos, icons, and inline SVGs), supplemented by
web archives when live endpoints are restricted---and reconstruct the full
page structure. Beyond static layout, we implement the interactive controls
required for end-to-end task completion, including multi-attribute filter
panels, date pickers, sorting menus, pagination, stateful submission forms, and
domain-specific components such as \texttt{Leaflet} maps and \texttt{Chart.js}
views.

\paragraph{Database grounding and multimodal completion.}
We populate each replica's seed database primarily from authentic structured
sources, including public APIs and data dumps (e.g., ESPN, Stack Exchange,
Hacker News, and OpenStreetMap), web crawl archives, and bibliographic catalogs.
When raw records lack full-text bodies, detailed item descriptions, or image
assets, we condition LLM text generation and image-generation models on the
authentic structured attributes to synthesize coherent articles, descriptions,
and context-matched thumbnails. All synthesized presentation fields are tagged
separately from the canonical attributes queried by task verifiers, ensuring
realistic visual and textual density without introducing ambiguity into
ground-truth grading.
\end{revblock}

\FloatBarrier
\subsection{Web Task Synthesis and Verifiable Contracts}
\label{app:web-task-synthesis}

\begin{figure}[!htbp]
    \centering
    \includegraphics[width=\linewidth]{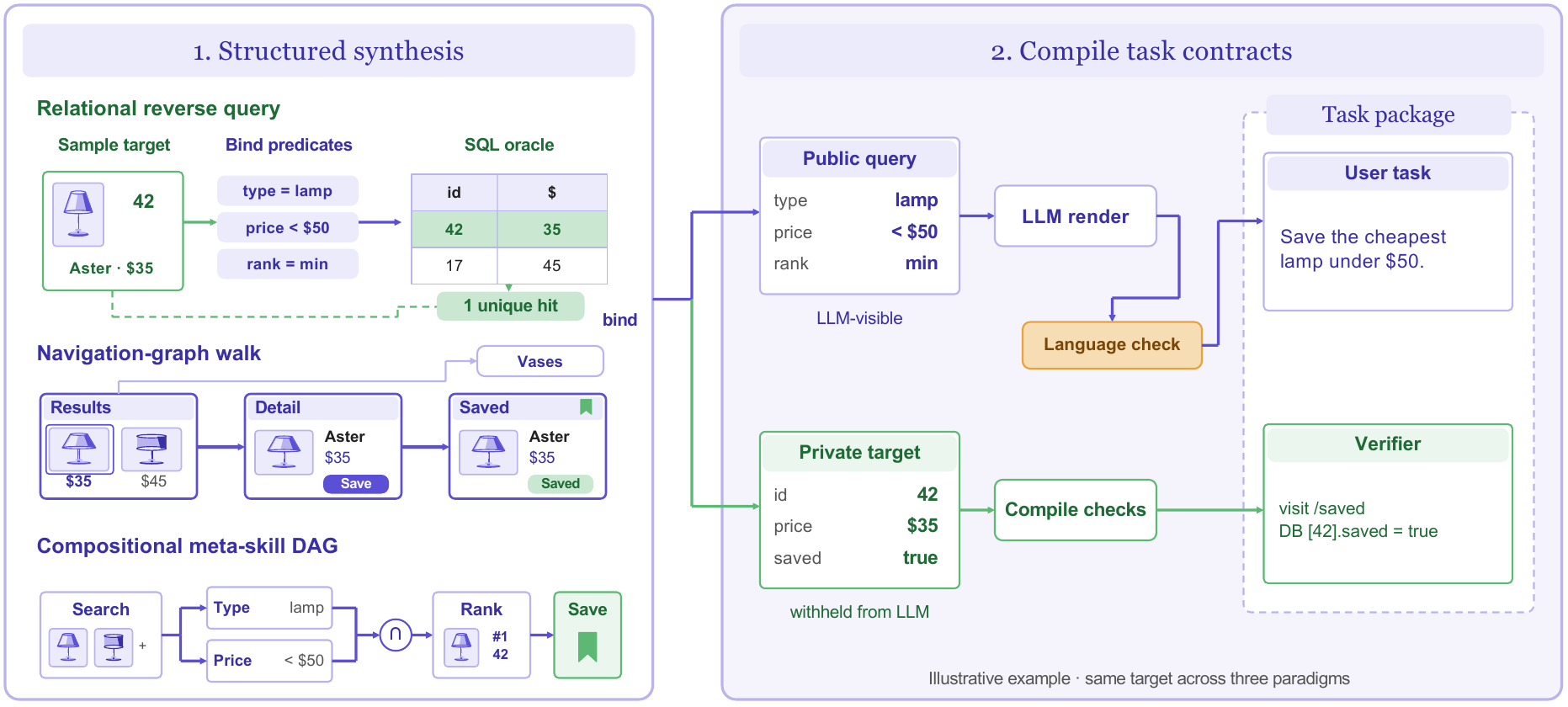}
    \caption{\rev{Structured task synthesis and contract compilation. The
    illustrative example links database predicates, a navigation path, and a
    compositional skill graph to a shared task specification. The instruction
    renderer receives only public query constraints; the verifier compiler uses
    the private target to check page visits and final database state.}}
    \label{fig:web-task-synthesis}
\end{figure}

\begin{revblock}
Building on the resettable replicas in Appendix~\ref{app:web-replicas}, we
formulate web task generation as a structured synthesis pipeline that pairs every
synthesized instruction with an executable verification contract
(Figure~\ref{fig:web-task-synthesis}). Each replica exposes its seed relational
database $\mathcal{D}_0$, a site navigation graph $\mathcal{G} = (\mathcal{V},
\mathcal{E})$ over page templates and UI transitions, and a set of stateful
backend operations. From these structured representations, we construct task
specifications through three complementary approaches:
\begin{enumerate}
    \item \textbf{Database-grounded query synthesis.} We sample target records
    and multi-attribute filter predicates directly from the seed database
    $\mathcal{D}_0$ and execute an SQL query over $\mathcal{D}_0$ to ensure that
    the resulting constraint set yields a unique, non-trivial answer before
    turning the predicates into a user task.
    \item \textbf{Navigation-path synthesis.} We sample executable multi-hop
    walks over the site's page-transition graph $\mathcal{G}$ (e.g., \emph{Home
    $\rightarrow$ Category Listing $\rightarrow$ Filtered Results $\rightarrow$
    Item Detail $\rightarrow$ Saved Collection}), converting each valid path
    into a goal-directed task whose required intermediate pages and terminal
    state are known by construction.
    \item \textbf{Compositional skill workflows.} Extending the workflow-first
    principle from Section~\ref{sec:synth} to browser environments, we
    abstract recurring web operations into typed \emph{meta-skills}---such as
    keyword search, multi-attribute filtering, comparative ranking, and stateful
    form submission---each equipped with explicit preconditions and
    postconditions. Composing compatible meta-skills into directed acyclic
    workflows across distinct entities in $\mathcal{D}_0$ produces diverse
    multi-step tasks that share a reusable verification template.
\end{enumerate}

\paragraph{Contract compilation and targeted curriculum expansion.}
Each task specification compiles along two separated branches
(Figure~\ref{fig:web-task-synthesis}, right). To render natural-language
instructions without leaking ground-truth answers into the prompt, we enforce
strict information asymmetry: the instruction-rendering LLM receives only the
public query constraints and allowed filter attributes, while the private target
records and oracle answers remain withheld; candidate prompts that fail
uniqueness or naturalness checks are discarded. In parallel, the verifier
compiler binds the private target records to deterministic checks over both the
execution trace (confirming required page visits and UI actions) and the final
outcome (checking the submitted answer and post-episode SQLite state). Following
the iterative feedback loop in Section~\ref{sec:taxonomy}, subsequent synthesis
rounds increase sampling weight on skill compositions where the student policy
underperforms---such as multi-constraint filtering or detail-page evidence
verification---supplying rule-verified browser and cross-application
trajectories for post-training.
\end{revblock}

\FloatBarrier
\subsection{Dataset Counting Definitions}

Tasks are deduplicated across 11 batches by domain and task identifier. A
task--file reference is a distinct path or filename extracted from a task's
metadata or text, including declared outputs. References are counted within
each task, with corresponding structured and textual declarations merged when
resolved.

Application coverage is multi-label, totaling 419,505 application--task
incidences across nine application categories. The 17,542
InfiniteWeb~\citep{zhang-etal-2026-infiniteweb} tasks are reported separately.
File categories are defined by extracted suffixes, including custom suffixes
and a no-extension category.

\section{Midtraining Explorations, Context Design, and Compute Accounting}
\label{app:training-quantities}

\subsection{GUI Midtraining Design Trade-Offs}
\label{app:midtraining-explorations}

\begin{figure}[!htbp]
    \centering
    \includegraphics[width=\linewidth]{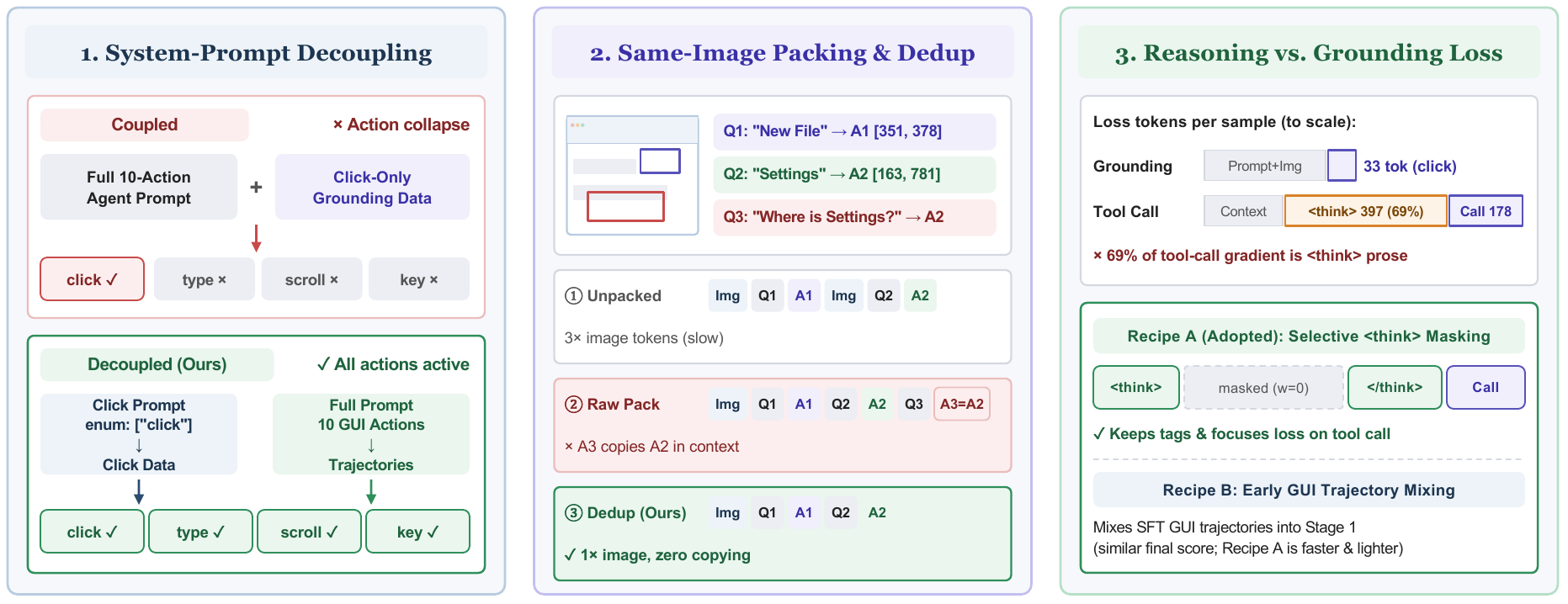}
    \caption{\rev{Three design choices in Stage~1 GUI-oriented midtraining:
    (1)~decoupling the click-only grounding prompt from the full agent schema to
    prevent action-space collapse, (2)~deduplicating repeated target elements
    within packed single-screenshot sequences to eliminate in-context coordinate
    copying, and (3)~selectively masking interior \texttt{<think>} tokens to
    balance reasoning format retention with grounding and tool-syntax learning.}}
    \label{fig:midtraining-tradeoffs}
\end{figure}

\begin{revblock}
Building on the Stage~1 overview in Section~\ref{sec:gui-midtraining},
Figure~\ref{fig:midtraining-tradeoffs} illustrates three design choices that
proved critical when jointly training screen grounding, structured tool calling,
and chain-of-thought reasoning.

\paragraph{Domain-targeted grounding and system-prompt decoupling.}
We curate our grounding pool from open-source GUI datasets, including
UGround~\citep{gou2025uground}, UI-E2I-Synth~\citep{liu2025uie2isynth},
OS-Atlas~\citep{wu2025osatlas}, MolmoWeb~\citep{molmoweb},
JEDI~\citep{xie2025osworldg}, and SeeClick~\citep{seeclick}, filtering for
720p--1080p desktop and browser screenshots and normalizing all bounding boxes
to center coordinates in $[0, 999]$ formatted as \texttt{computer\_use}
\texttt{click} tool calls. We pair these click-only examples with a minimal
grounding system prompt declaring only \texttt{"enum": ["click"]}
(Figure~\ref{fig:midtraining-tradeoffs}, left). Training click-only grounding
data under the full 10-action agent prompt causes the policy to associate the
full prompt with unconditional clicking and ignore \texttt{type},
\texttt{scroll}, or \texttt{keypress} downstream; decoupling the two prompts
transfers visual localization without collapsing the action space.

\paragraph{Target-level deduplication in same-screenshot packing.}
Packing multiple grounding queries for the same screenshot into a shared
sequence~\citep{aguvis} avoids repeated image encoding. However, open-source
grounding datasets often contain multiple paraphrased queries targeting the same
UI element on a screenshot (Figure~\ref{fig:midtraining-tradeoffs}, center).
When packed together, later turns achieve near-zero loss by copying the
coordinate pair emitted for an earlier turn ($A_k = A_j$) instead of attending
to the image. Retaining at most one query per distinct target location per
screenshot removes this shortcut and drives the steady grounding gains in
Figure~\ref{fig:midtraining-grounding}.

\paragraph{Reconciling non-reasoning grounding with reasoning tool calls.}
In packed sequences, a grounding example averages only 33 supervised action
tokens, whereas a reasoning-annotated tool-call example averages 575 supervised
tokens, 69\% of which belong to the \texttt{<think>} block
(Figure~\ref{fig:midtraining-tradeoffs}, right). Unmasked supervision lets
verbose reasoning dominate gradients and dilute localization, whereas stripping
\texttt{<think>} blocks causes the model to drop reasoning tags at test time. We
explored two complementary recipes:
\begin{enumerate}
    \item \textbf{Selective \texttt{<think>}-body masking on tool-call blends.}
    We mask the interior tokens of the \texttt{<think>} reasoning while
    supervising the opening and closing \texttt{<think>}...\texttt{</think>}
    delimiter tags and the \texttt{<tool\_call>}, preserving the think-then-act
    format at inference time while focusing gradients on tool syntax and
    coordinates.
    \item \textbf{Incorporating post-training GUI trajectories into midtraining.}
    We also explored mixing synthesized multi-step GUI trajectories directly
    into Stage~1 alongside grounding data, exposing the model early to the full
    downstream action space and step-level rationales.
\end{enumerate}
Both recipes reach comparable accuracy after downstream SFT; we adopt selective
\texttt{<think>}-body masking for the main pipeline because it keeps Stage~1
decoupled from trajectory synthesis, uses fewer tokens, and iterates faster.
\end{revblock}

\subsection{Single Step Cowork Bench}
\label{app:single-step-bench}

\begin{revblock}
Standard function-calling benchmarks such as BFCL evaluate text-only tool use
without visual inputs, standard GUI grounding benchmarks test only single-action
coordinate clicking, and end-to-end desktop benchmarks such as OSWorld require
downstream SFT and hours of multi-step VM rollouts per checkpoint. To jointly
evaluate visual localization, action-type selection, and reasoning format during
Stage~1 midtraining, we construct Single Step Cowork Bench from 342
representative interaction steps sampled from real desktop trajectories across
office suites (Calc, Writer, Impress), web browsing (Chrome), media and creative
tools (GIMP, VLC), development environments (VS Code), and system utilities.

Every example is evaluated under the full downstream agent system prompt and
spans eight GUI action primitives (\texttt{click}, \texttt{double\_click},
\texttt{triple\_click}, \texttt{drag}, \texttt{scroll}, \texttt{type},
\texttt{keypress}, and \texttt{terminate}). To separate direct localization from
intent-driven reasoning, each step is paired with both an \emph{explicit}
instruction (naming the target control directly, e.g., \emph{``Click the
Settings menu item in the popup menu''}) and an \emph{intent} instruction
(stating the local subgoal, e.g., \emph{``Open the VS Code settings page from
this menu''}). Grading jointly verifies (1)~valid \texttt{<think>} and
\texttt{<tool\_call>} syntax, (2)~exact action-type and argument match, and
(3)~spatial hit accuracy within the ground-truth bounding box. During Stage~1
development, this lightweight diagnostic surfaced failure modes invisible to
click-only grounding benchmarks---including action-space collapse onto
\texttt{click} under the full agent prompt, hallucinated placeholder function
names (\texttt{example\_function\_name}) copied from the schema, and
unterminated or missing \texttt{<think>} blocks on dense spreadsheet
views---allowing rapid recipe iteration before multi-step post-training.
\end{revblock}

\subsection{Context Window Design and Step-Level Reasoning Synthesis}
\label{app:context-relabeling}

\begin{figure}[!htbp]
    \centering
    \includegraphics[width=\linewidth]{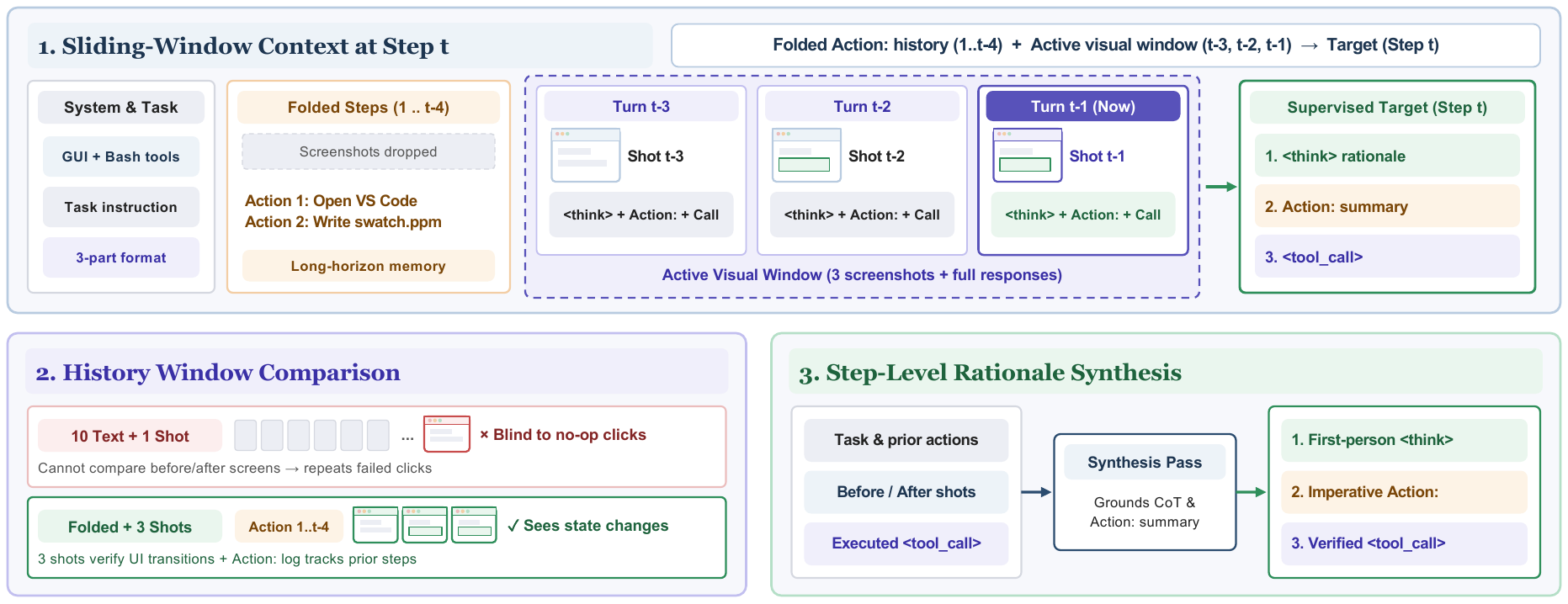}
    \caption{\rev{Sliding-window context design and step-level rationale
    synthesis: (1)~each turn conditions on folded imperative \texttt{Action:}
    summaries of steps $1\ldots t-4$ together with the three most recent visual
    turns ($t-3, t-2, t-1$), (2)~retaining three screenshots exposes whether
    prior actions altered the interface and prevents repetitive no-op click
    loops, and (3)~a reasoning synthesis pass grounds first-person
    \texttt{<think>} traces and \texttt{Action:} summaries on before/after
    screenshots and the executed tool call.}}
    \label{fig:context-window}
\end{figure}

\begin{revblock}
Long-horizon desktop trajectories face a fundamental context trade-off:
retaining every historical screenshot quickly exhausts the context budget,
whereas keeping only the current screenshot alongside text history leaves the
policy blind to whether its previous action actually changed the screen. In
early iterations with ten text turns and a single screenshot, a slightly
off-target click that left the interface unchanged frequently triggered
repetitive click loops (Figure~\ref{fig:context-window}, bottom left). To
resolve this, each turn-level training example uses an asymmetric sliding
window (Figure~\ref{fig:context-window}, top): it retains the three most recent
visual turns ($t-3, t-2, t-1$) with their screenshots and full assistant
responses so the model can directly verify before-and-after UI transitions,
while folding earlier steps ($1\ldots t-4$) into single-line imperative
\texttt{Action:} summaries that preserve a compact long-horizon record of
completed operations.

To supply consistent supervision for both in-window reasoning and out-of-window
history folding across heterogeneous trajectory sources, we run a step-level
reasoning synthesis pass conditioned on the task context, before/after
screenshots, and the executed tool call (Figure~\ref{fig:context-window},
bottom right). For every step, this pass produces a uniform three-part target:
(1)~a grounded first-person \texttt{<think>} rationale, (2)~a concise
imperative \texttt{Action:} summary that directly serves as the folded memory
entry once the turn slides past $t-3$, and (3)~the verified
\texttt{<tool\_call>}. Together with the asymmetric window, this uniform output
format turns the model's own generations into a self-sustaining working memory
across SFT and RL rollouts: recent turns provide visual and reasoning context
for local error recovery, while earlier \texttt{Action:} lines accumulate into a
compact execution log across long-horizon, interdependent tasks.
\end{revblock}

\subsection{Dataset Volumes and Compute Accounting}

For midtraining and SFT, nominal tokens equal the number of packed sequences
(751,346 for GUI-oriented midtraining, 983,462 for GUI agent SFT, 174,029 for
coding and terminal SFT, and 1,377,805 for hybrid CUA SFT) multiplied by their
32,768-token capacity, including padding. Nominal language-model compute is estimated as
$C\approx6N_{\mathrm{LM}}T$, with $N_{\mathrm{LM}}=4\times10^9$ and $T$ equal
to the two-epoch packed-token capacity. The estimates are
$1.182\times10^{21}$ FLOPs for midtraining, $3.988\times10^{21}$ for SFT, and
$2.618\times10^{21}$ for RL, totaling $7.787\times10^{21}$ FLOPs. Their shares
are 15.2\%, 51.2\%, and 33.6\%. The accounting covers the reported
language-model training stages, including image tokens in language-model
sequences; base pretraining, vision-encoder computation, and separate
attention/recomputation overhead terms are outside this scope.

Table~\ref{tab:historical-distributions} reports the mean, median, and P90 of
execution length and per-task and per-response token usage across training
stages (Section~\ref{sec:execution-diagnostics}), using the same counting
conventions as Figure~\ref{fig:historical-diagnostics}.

\begin{table}[!htbp]
    \centering
    \includegraphics[width=\linewidth]{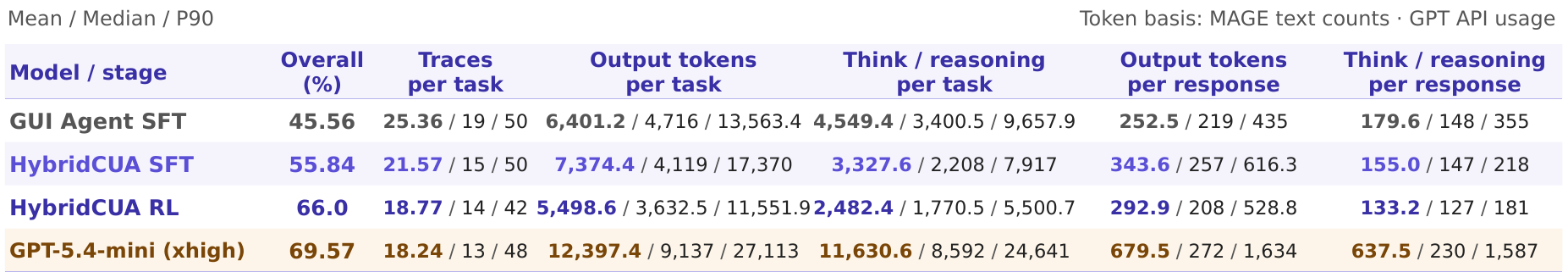}
    \caption{Execution and token statistics by training stage. Triples report
    mean, median, and P90; token-counting conventions are given in
    Section~\ref{sec:execution-diagnostics}.}
    \label{tab:historical-distributions}
\end{table}

\FloatBarrier
\input{sections/appendix_rl_infrastructure}

%% file: sections/appendix_rl_infrastructure.tex
\section{RL Infrastructure Details and Measurements}
\label{app:rl-system}

This appendix describes the RaaS implementation and deployment, followed by
workload, transport, update-timing, and policy-lag profiles. Each profile is
associated with its measurement run, update window, and aggregation rule.

\subsection{Rollout-as-a-Service Architecture}
\label{app:aaas-architecture}

The infrastructure integrates PoLar~\citep{polar}, Slime, Azure Relay, and
Azure Virtual Machine Scale Sets (VMSS). The training client uses Slime
adapters for desktop rollouts, turn-level samples, and data-parallel shards.
SGLang~\citep{sglang} serves model requests, while RaaS workers on Azure VMSS
run the resource manager and CUA runners. Routing and image preparation execute
on CPUs; inference and optimization execute on GPUs.

The relay exposes three logical communication channels:
\begin{itemize}[leftmargin=*,itemsep=2pt,topsep=3pt]
    \item \textbf{Control.} The training client submits work and queries
    task status and ready manifests from the resource manager.
    \item \textbf{Inference.} The harness sends prompts and screenshots to
    the inference service and receives generated tokens and log probabilities.
    \item \textbf{Bulk trajectories.} Stored compact episodes are delivered
    to the training adapters using deduplication and gzip compression.
\end{itemize}
The relay carries control, inference, and compact-trajectory traffic.
Trainer-to-inference weight synchronization and training collectives use
intra-cluster paths. External judge calls and artifact storage use their
respective service interfaces.

\subsection{Resource Management and Sandbox Execution}
\label{sec:infra-resource-manager}

The resource manager contains three components. The \emph{Task Dispatcher}
handles submission, queueing, dispatch, status, and ready manifests. The
\emph{Sandbox Scheduler} manages fleet capacity and sandbox allocation and
release, with separate admission limits for initialization, running, and
post-run processing. The \emph{Compact Episode JSON Store} gathers completed
trajectory records and supplies the bulk-transfer path. These components
separate task coordination and environment capacity from the model-serving
process.

Each host has separate concurrency caps for sandbox initialization, execution,
and post-run processing: 4 (\texttt{INIT}), 16 (\texttt{RUN}), and 8
(\texttt{POSTRUN}), respectively. Lifecycle-specific admission control manages
setup and post-processing alongside the 16 running-episode slots.

Each CUA runner contains four components. \emph{EnvRuntime} provides the
desktop environment, captures screenshots, and executes actions.
\emph{CUAHarness} maintains the stateful agent loop, issues model requests,
and dispatches GUI or terminal tools. \emph{Evaluator} produces the task's
evaluation, reward, and termination outcomes, using rule-based checks or
an external judge where required. \emph{TrajectoryBuilder} combines turn
traces and evaluator outcomes into compact episode JSON for transfer to the
training client.

\subsection{Asynchronous Rollouts and Trajectory Preparation}
\label{sec:infra-async-data}

The configured update batch contains 96 task groups with eight episode slots
per group, or 768 slots. The admission window spans four batches, with a
maximum configured policy lag of three optimizer updates. This organization
allows rollout collection and optimization to overlap while bounding the
policy-version gap of admitted data. Mixed-policy episodes are enabled.

TIS caps importance weights at 2.0. Admission is configured over a four-batch
window, with policy lag bounded at three optimizer updates
(Section~\ref{sec:rl}).

The data path keeps compact trajectories separate from expanded training
tensors. The rollout bridge and adapters construct per-turn samples and
data-parallel shards. Ray distributes compressed images, tokens, and metadata;
node-local CPU processing decodes and resizes the images and applies the
model's image processor before the tensors are consumed by the trainer GPUs.
Prefetching prepares trajectory data ahead of training consumption rather
than placing every preparation operation on the immediate update path.

Each turn retains up to three recent images. Compact episodes deduplicate
stored screenshots for transfer, and turn-level preparation materializes the
image inputs required by each training example.

\subsection{Deployment Scale and Runtime Profile}
\label{sec:infra-profile}

The CPU fleet comprises 70 Standard\_D64d\_v5 hosts, each with 64 vCPUs and
256 GiB of memory. Sixteen GUI slots per host provide 1,120 configured slots.
Each guest receives eight vCPUs and 4 GiB of memory, giving a 2:1 guest-to-host
vCPU allocation ratio. The reference GPU deployment uses 32 B200 GPUs, split
into 24 training and eight inference GPUs.

A deployment with 28 training GPUs and four inference GPUs reached a peak
occupancy of 1,120 running environments. The 24/8 split is the reference
configuration used in the architecture diagram.

\begin{figure}[H]
    \centering
    \includegraphics[width=0.92\linewidth]{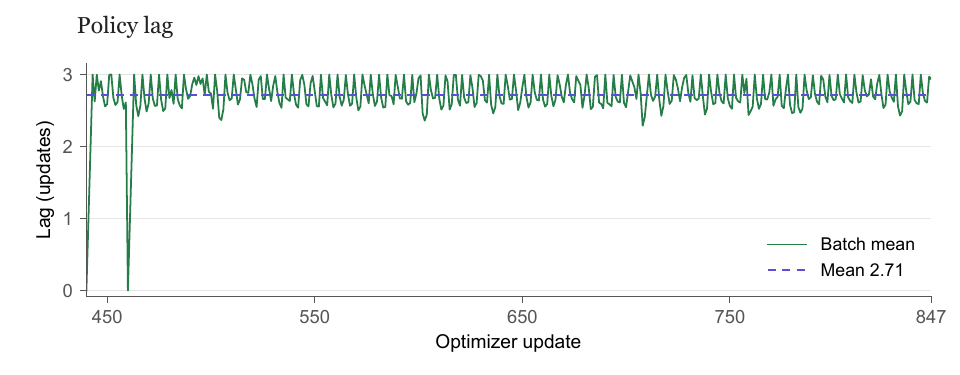}
    \caption{Runtime profile of policy lag.}
    \label{fig:runtime-profile}
\end{figure}

In the workload-profiling window, each update prepares 18,437.56 turn
sequences, 34,973.68 image occurrences, and 136.49 million language-model
tokens on average. Compressed episode payloads average 2.44 GB (2.27 GiB) per
update. Node-local preprocessing materializes 408.25 GiB of CPU pixel/grid
tensors across 24 trainer ranks.

Across 11 post-initialization updates, the mean update time is 570.3 seconds,
corresponding to 6.31 updates per hour in that window. The mean of 408
update-level policy-lag measurements is 2.71 updates, with observed batch means
ranging from 0.0 to 3.0.

\subsection{Workload and Transport Accounting}

Turn-sequence counts include all prepared samples, including zero-mask
sequences. Image-occurrence counts include screenshot inputs repeated across
turn histories. The context retains up to three recent images per turn.

Bulk volume measures successful compressed response payloads for consumed
rollout groups, averaging 2.44 GB (2.27 GiB) per update. Protocol headers and
retransmissions are outside this payload measure. Prepared CPU pixel/grid
tensors average 408.25 GiB per update across 24 trainer ranks.

\subsection{Policy-Lag Aggregation}

For a consumed turn-level training sample $j$, the logged lag is
\begin{equation}
    \ell_j = \max(0, a_j-v_j), \qquad
    L_u = \frac{1}{|\mathcal{B}_u|}
          \sum_{j\in\mathcal{B}_u}\ell_j,
    \label{eq:policy-lag}
\end{equation}
Here $a_j$ denotes the accepted rollout index, $v_j$ the behavior-policy
version index, and $\mathcal{B}_u$ the turn samples consumed at update $u$.
Version indices follow the common numbering convention used to compute the
policy-lag statistic.

The reported policy lag of 2.71 updates is the arithmetic mean of the 408
update-level means $L_u$. These means range from 0.0 to 3.0 updates, within the
configured three-update lag bound.